# Fast Reinforcement Learning for Energy-Efficient Wireless Communications

Nicholas Mastronarde and Mihaela van der Schaar

**Abstract -** We consider the problem of energy-efficient point-to-point transmission of delay-sensitive data (e.g. multimedia data) over a fading channel. Existing research on this topic utilizes either physical-layer centric solutions, namely power-control and adaptive modulation and coding (AMC), or system-level solutions based on dynamic power management (DPM); however, there is currently no rigorous and unified framework for simultaneously utilizing both physical-layer centric and system-level techniques to achieve the minimum possible energy consumption, under delay constraints, in the presence of stochastic and a priori unknown traffic and channel conditions. In this report, we propose such a framework. We formulate the stochastic optimization problem as a Markov decision process (MDP) and solve it online using reinforcement learning. The advantages of the proposed online method are that (i) it does not require a priori knowledge of the traffic arrival and channel statistics to determine the jointly optimal power-control, AMC, and DPM policies; (ii) it exploits partial information about the system so that less information needs to be learned than when using conventional reinforcement learning algorithms; and (iii) it obviates the need for action exploration, which severely limits the adaptation speed and run-time performance of conventional reinforcement learning algorithms. Our results show that the proposed learning algorithms can converge up to two orders of magnitude faster than a state-of-the-art learning algorithm for physical layer power-control and up to three orders of magnitude faster than conventional reinforcement learning algorithms.

N. Mastronarde is with the Department of Electrical Engineering, State University of New York at Buffalo, Buffalo, NY 14260, USA (e-mail: nmastron@buffalo.edu). This work was done while he was at the University of California at Los Angeles (UCLA), Los Angeles, CA 90095-1594, USA.
M. van der Schaar is with the Department of Electrical Engineering, University of California at Los Angeles (UCLA), Los Angeles, CA 90095-1594, USA (e-mail: mihaela@ee.ucla.edu).
This work was supported in part by Sanyo, Japan, and NSF grant no. 0831549.

I. INTRODUCTION

Delay-sensitive communication systems often operate in dynamic environments where they experience time-varying channel conditions (e.g. fading channel) and dynamic traffic loads (e.g. variable bit-rate). In such systems, the primary concern has typically been the reliable delivery of data to the receiver within a tolerable delay. Increasingly, however, battery-operated mobile devices are becoming the primary means by which people consume, author, and share delay-sensitive content (e.g. real-time streaming of multimedia data, videoconferencing, gaming etc.). Consequently, energy-efficiency is becoming an increasingly important design consideration. To balance the competing requirements of energy-efficiency and low delay, fast learning algorithms are needed to quickly adapt the transmission decisions to the time-varying and a priori unknown traffic and channel conditions.

Existing research that addresses the problem of energy-efficient wireless communications can be roughly divided into two categories: physical (PHY) layer-centric solutions such as power-control and adaptive modulation and coding (AMC); and system-centric solutions such as dynamic power management (DPM).[1] Although these techniques differ significantly, they can all be used to tradeoff delay and energy to increase the lifetime of battery-operated mobile devices [2] [3] [8].

**PHY-centric solutions:** A plethora of existing PHY-centric solutions focus on optimal single-user power-control (also known as *minimum energy transmission* or *optimal scheduling*) with the goal of minimizing transmission power subject to queuing delay constraints (e.g. [1] [2]). It is well known that transmitting with more power in good channel states, and with less power in poor channel states, maximizes throughput under a fixed energy budget. For this reason, [1] and [2] use dynamic programming, coupled with a stochastic fading channel model, to determine the optimal power-control policy. Unfortunately, the aforementioned solutions require statistical knowledge of the underlying dynamics (i.e. the channel state and traffic distributions), which is typically not available in practice. When this information is not available, only heuristic solutions, which cannot guarantee optimal performance, are provided. For example, in [2], a heuristic power-control policy is proposed that is only optimal for asymptotically large buffer delays. Moreover, [1] and [2] are information-theoretic in nature, so they ignore transmission errors, which commonly occur in practical wireless transmission scenarios.

---

[1] DPM enables system components such as the wireless network card to be put into low-power states when they are not needed [3]-[6]. We discuss this in more detail in Section II.B.

Other PHY-centric solutions are based on adaptive modulation, adaptive coding, or AMC. Although these techniques are typically used to tradeoff throughput and error-robustness in fading channels [22], they can also be exploited to tradeoff delay and energy as in [8], where they are referred to as dynamic modulation scaling, dynamic code scaling, and dynamic modulation-code scaling, respectively. Offline and online techniques for determining optimal scaling policies are proposed in [8], however, these techniques cannot be extended in a manner that tightly integrates PHY-centric and system-level power management techniques.

Although PHY-centric solutions are effective at minimizing transmission power, they ignore the fact that it costs power to keep the wireless card on and ready to transmit; therefore, a significant amount of power can be wasted even when there are no packets being transmitted. System-level solutions address this problem.

**System-level solutions:** System-level solutions rely on DPM, which enables system components such as the wireless network card to be put into low-power states when they are not needed [3]-[6]. A lot of work has been done on DPM, ranging from rigorous work based on the theory of Markov decision processes [3]-[5], to low-complexity work based on heuristics [6]. In [3], the optimal DPM policy is determined offline under the assumption that the traffic arrival distribution is known a priori. In [5], an online approach using maximum likelihood estimation to estimate the traffic arrival distribution is proposed. This approach requires using the complex value iteration algorithm to update the DPM policy to reflect the current estimate of the traffic distribution. In [4], a supervised learning approach is taken to avoid the complex value iteration algorithm. However, this approach incurs large memory overheads because it requires many policies to be computed offline and stored in memory to facilitate the online decision making process. Unfortunately, due to their high computational and memory complexity, the aforementioned online approaches are impractical for optimizing more complex, resource-constrained, and delay-sensitive communication systems.

Our contributions are as follows:

- **Unified power management framework:** We propose a rigorous and unified framework, based on Markov decision processes (MDPs) and reinforcement learning (RL), for simultaneously utilizing both PHY-centric and system-level techniques to achieve the minimum possible energy consumption, under delay constraints, in the presence of stochastic traffic and channel conditions. This is in contrast to existing work that only utilizes power-control [1] [2], AMC [8], or DPM [3]-[6] to manage power.[2]

- **Generalized post-decision state:** We propose a decomposition of the (offline) value iteration and (online) RL algorithms based on factoring the system's dynamics into a priori known and a priori unknown components. This is achieved by generalizing the concept of a *post-decision state* (PDS, or *afterstate* [11]), which is an intermediate state that occurs after the known dynamics take place but before the unknown dynamics take place. Similar decompositions have been used for modeling games such as tic-tac-toe and backgammon [11], for dynamic channel allocation in cellular telephone systems [11], and for delay-constrained scheduling over fading channels (i.e. power-control) [12] [16]. However, we provide a more general formulation of the PDS concept than existing literature. Specifically, existing literature introduces the PDS concept for very specific problem settings, where the PDS is a deterministic function of the current state and action, and where the cost function is assumed to be known. In general, however, the PDS can be a non-deterministic function of the current state and action, and it can be used in any MDP in which it is possible to factor the transition probability and cost functions into known and unknown components. The advantages of the proposed PDS learning algorithm are that it exploits partial information about the system, so that less information needs to be learned than when using conventional RL algorithms and, under certain conditions, it obviates the need for action exploration, which severely limits the adaptation speed and run-time performance of conventional RL algorithms.

- **Virtual experience:** Unlike existing literature, we take advantage of the fact that the unknown dynamics are independent of certain components of the system's state. We exploit this property to perform a batch update on multiple PDSs in each time slot. We refer to this batch update as virtual experience learning. Prior to this work, it was believed that the PDS learning algorithm must necessarily be performed one state at a time because one learns only about the current state being observed, and can, therefore, update only the corresponding component [12]. Importantly, our experimental results illustrate that virtual experience can reduce the convergence time of learning by up to two orders of magnitude compared to "one state at a time" PDS learning.

---

[2] More accurately, prior research on power-control implicitly considers adaptive coding because it assumes that optimal codes, which achieve information-theoretic capacity, are used.

- **Application of reinforcement learning to dynamic power management (DPM):** We believe that this report is the first to apply RL to the DPM problem. This is non-trivial because naïve application of RL can lead to very poor performance. This is because conventional RL solutions (e.g. Q-learning) require frequent action exploration, which lead to significant power and delay penalties when a suboptimal power management action is taken. Using PDS learning in conjunction with virtual experience learning, however, obviates the need for action exploration and allows the algorithm to quickly adapt to the dynamics and learn the optimal policy.

The primary limitation of our algorithm is that, although it generally performs well, it is not provably optimal under non-Markovian dynamics. In fact, when the traffic is highly time-varying and the Markovian assumption is violated, the proposed solution can occasionally perform worse than a simple threshold policy. This phenomenon has been observed in prior literature on dynamic power management. For example, it has been shown in [3] that simple timeout policies can occasionally outperform MDP-based policies in non-Markovian environments.

The remainder of the report is organized as follows. In Section II, we introduce the system model and assumptions. In Section III, we describe the transmission buffer and traffic models. In Section IV, we formulate the power management problem as an MDP. In Section V, we discuss how to solve the problem online using RL. In Section VI, we present our simulation results. Section VII concludes the report.

## II. Preliminaries

In this section, we introduce the time-slotted system model used in this report. We assume that time is slotted into discrete-time intervals of length $\Delta t$ such that the $n$th time slot is defined as the time interval $[n\Delta t, (n+1)\Delta t)$. Transmission and power management decisions are made at the beginning of each interval and the system's state is assumed to be constant throughout each interval. Figure 1 illustrates the considered wireless transmission system. We describe the deployed PHY-centric power management techniques in Section II.A and the deployed system-level power management technique in Section II.B. We describe the transmission buffer model in Section III. List of notation and abbreviations are provided in Table 1 and

Table **2**, respectively.

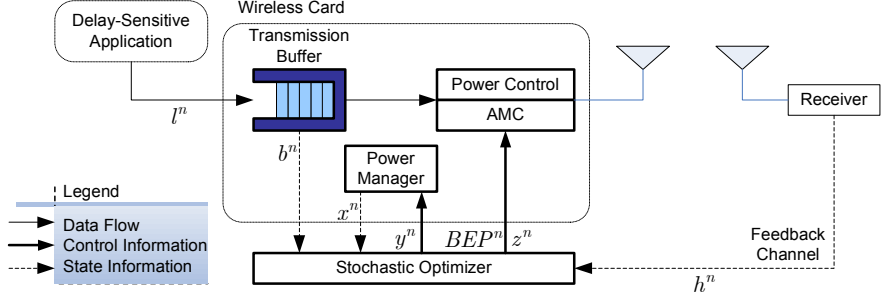

**Figure 1. Wireless transmission system. The components outlined in bold are the focus of this report.**

**Table 1. List of notation.**

| Notation | Meaning |
|---|---|
| $n$, $\Delta t$ | Time slot index, time slot duration |
| $b$, $h$, $x$, $s$ | Buffer state, channel state, power management state, joint state |
| $\tilde{b}$, $\tilde{h}$, $\tilde{x}$, $\tilde{s}$ | Post-decision states |
| $BEP$, $y$, $z$, $a$ | Bit-error probability, power management action, packet throughput, joint action |
| $\rho([h,x], BEP, y, z)$, $P_{\text{tx}}(h^n, BEP^n, z^n)$, $P_{\text{on}}$, $P_{\text{off}}$, $P_{\text{tr}}$, $\theta$ | Power cost, transmission power, "on" power, "off" power, transition power, power management transition probability |
| $g([b,x], BEP, y, z)$, $\eta$, $B$, $\delta$ | Buffer cost, dropped packet penalty, maximum buffer size, buffer delay constraint |
| $L$, $l$, $f$, $PLR$ | Packet size, packet arrivals, packet goodput, packet loss rate |
| s_on, s_off | Switch on / stay on, switch off / stay off |
| $\gamma$ | Discount factor |
| $\bar{P}^\pi(s)$, $\bar{D}^\pi(s)$, $L^{\pi,\mu}(s)$ | Discounted average power, discounted average delay, discounted average Lagrangian cost |
| $\mu$, $\Lambda$ | Lagrange multiplier, projection operator |
| $c^\mu(s,a)$, $c_{\text{k}}(s^n, a)$, $c_{\text{u}}(\tilde{s}, a)$ | Lagrangian cost, known cost, unknown cost |
| $V(s)$, $\tilde{V}(\tilde{s})$, $Q(s,a)$, $\pi$ | State-value function, post-decision state-value function, action-value function, policy |
| $\alpha$, $\varepsilon$, $\beta$ | Value function learning rate, exploration rate, Lagrange multiplier learning rate |
| $p^b(b' \mid [b,h,x], BEP, y, z)$, $p^h(h' \mid h)$, $p^x(x' \mid x, y)$, $p_{\text{k}}(\tilde{s} \mid s, a)$, $p_{\text{u}}(s' \mid \tilde{s}, a)$ | Buffer, channel, power management, known, and unknown transition probabilities |
| $\sigma$, $\tilde{\sigma}^n$, $\sum(\tilde{\sigma}^n)$ | Experience tuple, virtual experience tuple, set of virtual experience tuples |
| $N_0$, $W$, $T_s$ | Noise power spectral density, bandwidth, symbol duration |
| $e^{n,\pi^*}$, $p^*(s)$, $n^\epsilon_{\text{converge}}$ | Weighted error metric, stationary distribution under the optimal policy, $\epsilon$-convergence time |

**Table 2. List of abbreviations.**

| Abbreviations | Meaning |
|---|---|
| AMC | Adaptive modulation and coding |
| BEP | Bit-error probability |
| DPM | Dynamic power management |
| ET | Experience tuple |
| i.i.d. | Independent and identically distributed |
| MDP | Markov decision process |
| PDS | Post-decision state |
| PHY | Physical layer |
| PLR | Packet loss rate |
| RL | Reinforcement learning |
| SNR | Signal-to-noise ratio |

*A. Physical layer: adaptive modulation and power-control*

We consider the case of a frequency non-selective channel, where $h^n \in \mathcal{H}$ denotes the fading coefficient over the point-to-point link (i.e., from the transmitter to the receiver) in time slot $n$ as illustrated in Figure 1. As in [1], [2], [12], and [13], we assume that the set of channel states $\mathcal{H}$ is discrete and finite, and that the channel state is constant for the duration of a time slot, that it can be estimated perfectly, and that the sequence of channel states $\{h^n \in \mathcal{H} : n = 0, 1, \ldots\}$ can be modeled as a Markov chain with transition probabilities $p^h(h' \mid h)$.

The physical layer is assumed to be a single-carrier single-input single-output (SISO) system with fixed symbol rate of $1 / T_s$ (symbols per second). The transmitter sends at a data rate $\beta^n / T_s$ (bits/s) to the receiver, where $\beta^n \geq 1$ is the number of bits per symbol determined by the modulation scheme in the AMC component shown in Figure 1. All packets are assumed to have packet length $L$ (bits) and the symbol period $T_s$ is fixed.

The proposed framework can be applied to any modulation and coding schemes. Our only assumptions are that the bit-error probability (BEP) at the output of the maximum likelihood detector of the receiver, denoted by $BEP^n$, and the transmission power, denoted by $P_{\text{tx}}^n$, can be expressed as

$$BEP^n = BEP(h^n, P_{\text{tx}}^n, z^n) \text{ and} \tag{1}$$

$$P_{\text{tx}}^n = P_{\text{tx}}(h^n, BEP^n, z^n), \tag{2}$$

where $z^n$ is the *packet throughput* in packets per time slot. Assuming independent bit-errors, the packet loss rate (PLR) for a packet of size $L$ can be easily computed from the BEP as

$$PLR^n = 1 - \left(1 - BEP^n\right)^L.$$

We will use the packet throughput as a decision variable in our optimization problem as shown in Figure 1. To achieve a packet throughput of $z^n \in \mathcal{Z}$ (i.e., to transmit $Lz^n$ bits in $\Delta t$ seconds), the modulation scheme must set the number of bits per symbol to

$$\beta^n = \lceil z^n L T_s / \Delta t \rceil,$$

where $\lceil x \rceil$ denotes the smallest integer that is greater than $x$. Given the packet throughput, which determines the number of bits per symbol, and channel state $h^n$, we are then free to either select (i) the transmission power $P_{\text{tx}}^n$, from which we can determine the BEP using (1), or select (ii) the BEP, from which we can determine the transmission power using (2). We select the BEP as our decision variable, as shown in Figure 1, but the proposed framework can also be applied if we select the transmission power.

For illustration, throughout this report we assume that (1) and (2) are known; however, the proposed learning framework is general and can be directly extended to the case when (1) and (2) are unknown a priori. We also select a fixed coding scheme for simplicity; however, the proposed framework can still be applied using adaptive modulation and coding with an appropriate modification to the BEP and transmission power functions. Note that if both modulation and coding can be adapted, then (1) and (2) are often unattainable analytically. Thus, if (1) and (2) are assumed to be known, it is necessary to resort to models based on fitting parameters to measurements as in [19]-[21].

## B. System-level: dynamic power management model

In addition to adaptive modulation and power-control, which determine the active transmission power, we assume that the power manager illustrated in Figure 1 can put the wireless card into low power states to reduce the power consumed by the wireless card's electronic circuits. Specifically, the wireless card can be in one of two power management states in the set $\mathcal{X} = \{\text{on}, \text{off}\}$ and can be switched on and off using one of the two power management actions in the set $\mathcal{Y} = \{\text{s\_on}, \text{s\_off}\}$.[3] As in [3], the notation s_on and s_off should be read as "switch on" and "switch off," respectively.

If the channel is in state $h$, the wireless card is in state $x$, the maximum BEP is $BEP^n$, the power

---

[3] This can be easily extended to multiple power management states as in [4], but we only consider two for ease of exposition.

management action is $y$, and the packet throughput is $z$, then the required power is

$$\rho([h,x],BEP,y,z) = \begin{cases} [P_{\text{on}} + P_{\text{tx}}(h^n, BEP^n, z^n)], & \text{if } x = \text{on}, y = \text{s\_on} \\ P_{\text{off}}, & \text{if } x = \text{off}, y = \text{s\_off} \\ P_{\text{tr}}, & \text{otherwise}, \end{cases} \quad (3)$$

where $P_{\text{tx}}$ (watts) is the transmission power defined in (2), $P_{\text{on}}$ and $P_{\text{off}}$ (watts) are the power consumed by the wireless card in the "on" and "off" states, respectively, and $P_{\text{tr}}$ (watts) is the power consumed when it transitions from "on" to "off" or from "off" to "on". Similar to [3], we assume that $P_{\text{tr}} \geq P_{\text{on}} > P_{\text{off}} \geq 0$ such that there is a large penalty for switching between power states, but remaining in the "off" state consumes less power than remaining in the "on" state.[4]

As in [3], we model the sequence of power management states $\{x^n \in \mathcal{X} : n = 0,1,...\}$ as a controlled Markov chain with transition probabilities $p^x(x' \mid x,y)$. Let $\mathbf{P}^x(y)$ denote the transition probability matrix conditioned on the power management action $y$, such that $\mathbf{P}^x(y) = [p^x(x' \mid x,y)]_{x,x'}$. Due to the high level of abstraction of the model, it is possible that there is a non-deterministic delay associated with the power management state transition such that

$$\mathbf{P}^x(\text{s\_on}) = \begin{array}{c} \text{on} \\ \text{off} \end{array} \begin{pmatrix} \overset{\text{on}}{1} & \overset{\text{off}}{0} \\ \theta & 1-\theta \end{pmatrix}$$

$$\mathbf{P}^x(\text{s\_off}) = \begin{array}{c} \text{on} \\ \text{off} \end{array} \begin{pmatrix} \overset{\text{on}}{1-\theta} & \overset{\text{off}}{\theta} \\ 0 & 1 \end{pmatrix} \quad (4)$$

where the row and column labels represent the current ($x^n$) and next ($x^{n+1}$) power management states, respectively, and $\theta \in (0,1]$ (resp. $1-\theta$) is the probability of a successful (resp. an unsuccessful) power state transition. For simplicity of exposition, we will assume that the power state transition is deterministic (i.e. $\theta = 1$); however, our framework applies to the case with $\theta < 1$ (with $\theta$ known or unknown). We note that the packet throughput $z$ can be non-zero only if $x = \text{on}$ and $y = \text{s\_on}$; otherwise, the packet throughput $z = 0$.

---

[4] In [32], the authors find that switching from the doze mode ("off") to the idle mode ("on") consumes the same power as being in the idle mode. However, as noted in [33], different manufacturers have different implementations of the power-save mode, leading to different switching power penalties, which our framework can be easily adapted to cope with.

## III. TRANSMISSION BUFFER AND TRAFFIC MODEL

We assume that the transmission buffer is a first-in first-out queue. As illustrated in Figure 1, the source injects $l^n$ packets into the transmission buffer in each time slot, where $l^n$ has distribution $p^l(l)$. Each packet is of size $L$ bits and the arrival process $\{l^n : n = 0,1,...\}$ is assumed to be independent and identically distributed (i.i.d) with respect to the time slot index $n$. The arriving packets are stored in a finite-length buffer, which can hold a maximum of $B$ packets. The buffer state $b \in \mathcal{B} = \{0,1,...,B\}$ evolves recursively as follows:

$$b^0 = b_{\text{init}}$$
$$b^{n+1} = \min(b^n - f^n(BEP^n, z^n) + l^n, B), \tag{5}$$

where $b_{\text{init}}$ is the initial buffer state and $f^n(BEP^n, z^n) \leq z^n$ is the *packet goodput* in packets per time slot (i.e. the number of packets transmitted without error), which depends on the throughput $z^n \in \{0,...,\min(b^n, z_{\max})\} = \mathcal{Z}(b^n)$ and the BEP $BEP^n \in \mathcal{E}$.[5] Recall that the packet throughput $z$, and therefore the packet goodput $f$, can be non-zero only if $x = \text{on}$ and $y = \text{s\_on}$; otherwise, they are both zero (i.e. $z = f = 0$). Also, note that the packets arriving in time slot $n$ cannot be transmitted until time slot $n+1$ and that any unsuccessfully transmitted packets stay in the transmission buffer for later (re)transmission. For notational simplicity, we will omit the arguments of the packet goodput and instead write $f^n = f^n(BEP^n, z^n)$. Assuming independent packet losses, $f^n$ is governed by a binomial distribution, i.e.

$$p^f(f^n \mid BEP^n, z^n) = \text{bin}(z^n, 1 - PLR^n), \tag{6}$$

and has expectation $E[f^n] = (1 - PLR^n)z^n$. We refer to $p^f(f^n \mid BEP^n, z^n)$ as the goodput distribution.

Based on the buffer recursion in (5), the arrival distribution $p^l(l)$, and the goodput distribution $p^f(f \mid BEP, z)$, the sequence of buffer states $\{b^n : n = 0,1,...\}$ can be modeled as a controlled Markov chain with transition probabilities

---

[5] Note that, by enforcing $z^n \leq b^n$, we do not have to include underflow conditions in the buffer evolution equation (5).

$$p^b\left(b' \mid [b,h,x], BEP, y, z\right) = \begin{cases} \sum_{f=0}^{z} p^l\left(b' - [b-f]\right) p^f\left(f \mid BEP, z\right), & \text{if } b' < B \\ \sum_{f=0}^{z} \sum_{l=B-[b-f]}^{\infty} p^l(l) p^f\left(f \mid BEP, z\right), & \text{if } b' = B. \end{cases} \quad (7)$$

Eq. (7) is similar to the buffer state transition probability function in [9], except that the model in [9] assumes that the packet goodput $f$ is equal to the packet throughput $z$ (i.e. there are no packet losses).

We also define a *buffer cost* to reward the system for minimizing queuing delays, thereby protecting against overflows that may result from a sudden increase in transmission delay due to a bad fading channel or traffic burst. Formally, we define the buffer cost as the expected sum of the *holding cost* and *overflow cost* with respect to the arrival and goodput distributions: i.e.,

$$g([b,x], BEP, y, z) = \sum_{l=0}^{\infty} \sum_{f=0}^{z} p^l(l) p^f(f \mid BEP, z) \left\{ \underbrace{[b-f]}_{\text{holding cost}} + \underbrace{\eta \max([b-f]+l-B, 0)}_{\text{overflow cost}} \right\}. (8)$$

In (8), the holding cost represents the number of packets that were in the buffer at the beginning of the time slot, but were not transmitted (and hence must be held in the buffer to be transmitted in a future time slot). If the buffer is stable (i.e. there are no buffer overflows), then the holding cost is proportional to the queuing delay by Little's theorem [10]. If the buffer is not stable (i.e. there is a finite probability of overflow such that Little's theorem does not apply), then the overflow cost imposes a penalty of $\eta$ for each dropped packet. We describe how to define $\eta$ at the end of Section IV, after introducing the formal optimization problem, because its definition requires a parameter used in the optimization.

## IV. WIRELESS POWER MANAGEMENT PROBLEM FORMULATION

### A. Formulation as a constrained Markov decision process

In this section, we formulate the wireless power management problem as a constrained MDP. We define the joint state (hereafter *state*) of the system as a vector containing the buffer state $b$, channel state $h$, and power management state $x$, i.e. $s \triangleq (b,h,x) \in \mathcal{S}$. We also define the joint action (hereafter *action*) as a vector containing the BEP $BEP$, power management action $y$, and packet throughput $z$, i.e. $a \triangleq (BEP, y, z) \in \mathcal{A}$. The sequence of states $\{s^n : n = 0, 1, ...\}$ can be modeled as a controlled Markov chain with transition probabilities that can be determined from the conditionally independent buffer state, channel state, and power management state transitions as follows:

$$p(s' \mid s,a) = p^b\left(b' \mid [b,h,x], BEP, y, z\right) p^h\left(h' \mid h\right) p^x\left(x' \mid x, y\right). \tag{9}$$

The objective of the wireless power management problem is to minimize the *infinite horizon discounted power cost* subject to a constraint on the *infinite horizon discounted delay*. We describe below why we use a *discounted* criterion instead of, for example, a time average criterion. Let $\pi : S \mapsto \mathcal{A}$ denote a stationary *policy* mapping states to actions such that $a = \pi(s)$, and let $\Phi$ denote the set of all possible stationary policies. Starting from state $s$ and following policy $\pi \in \Phi$, the expected discounted power cost and expected discounted delay are defined as

$$\bar{P}^\pi(s) = E\left[\sum_{n=0}^{\infty} (\gamma)^n \rho(s^n, \pi(s^n)) \mid s^0 = s\right], \text{ and} \tag{10}$$

$$\bar{D}^\pi(s) = E\left[\sum_{n=0}^{\infty} (\gamma)^n g(s^n, \pi(s^n)) \mid s^0 = s\right], \tag{11}$$

where the expectation is over the sequence of states $\{s^n : n = 0,1,...\}$, $\gamma \in [0,1)$ is the *discount factor*, and $(\gamma)^n$ denotes the discount factor to the $n$th power. Formally, the objective of the constrained wireless power management problem is

$$\text{Minimize } \bar{P}^\pi(s) \text{ subject to } \bar{D}^\pi(s) \leq \delta, \forall s \in \mathcal{S}, \tag{12}$$

where $\delta$ is the discounted buffer cost constraint. Note that, although $\delta$ is technically a constraint on (11) (i.e. the discounted buffer cost, which includes the overflow cost), we will often refer to it as the "delay constraint" or "holding cost constraint."[6] This is appropriate because feasible solutions to (12) typically incur negligible overflow costs as demonstrated by our simulations in Section VI. A necessary condition for the existence of a feasible solution to (12) is that there exists at least one transmission action $z \in \mathcal{Z}$ that is larger than the expected packet arrival rate, i.e. there must exist a $z > \sum_{l=0}^{\infty} p_l(l) \cdot l$.

It turns out that the discounted cost criterion is mathematically more convenient than an average cost criterion due to the structure of the considered MDP. Specifically, unlike [9] [12] [13], we consider an MDP that is *multichain* rather than *unichain*.[7] Importantly, it is multichain because we consider DPM.

---

[6] Note that we impose a constraint on the (discounted) average delay. This is different from a hard delay constraint, which would guarantee that no single packet is delayed by more than $\delta$ seconds. Unfortunately, due to the stochastic nature of the problem, it is not possible to guarantee a hard delay constraint.

[7] An MDP is unichain if the transition matrix corresponding to *every* deterministic stationary policy $\pi : S \mapsto \mathcal{A}$ is unichain, that is, it consists of a single recurrent class plus a possibly empty set of transient states [27]. Meanwhile, an MDP is multichain if the transition matrix corresponding to *at least one* stationary policy contains two or more closed irreducible classes [27].

This can be seen by imagining a DPM policy that keeps the wireless card "off" in the "off" state "on" in the "on" state, thereby inducing two closed irreducible recurrent classes of states. As discussed in Chapters 8 and 9 of [27], for unichain MDPs, a single optimality equation is enough to characterize the optimal policy under an average reward objective. For multichain MDPs, however, a single optimality equation may not be sufficient to characterize the optimal policy. Consequently, algorithms for multichain models are more complex, and the theory is less complete, than for unichain models. We chose to use the discounted criterion because it only requires a single optimality equation regardless of the chain structure of the MDP

In addition to mathematical convenience, there are several intuitive justifications for using discounted objectives and constraints, which weigh immediate costs more heavily than expected future costs. First, in practice, traffic and channel dynamics are only "stationary" over short time intervals; consequently, forecasts of costs that a policy will incur far in the future are unreliable (because they are based on past experience which does not have the same statistics as future experience), so expected future costs should be discounted when making decisions in each time slot. Second, we may indeed want to optimize the time average cost, but because the application's lifetime is not known a priori (e.g. a video conference may end abruptly), we assume that the application will end with probability $1 - \gamma$ in any time slot (i.e., in any state, with probability $1 - \gamma$, the MDP will transition to a zero cost absorbing state). This second interpretation is thoroughly described in [3].

Now that we are familiar with the optimization objective and constraint, we can describe how to define the per-packet overflow penalty $\eta$ used in the buffer cost function in (8). $\eta$ must be large enough to ensure that it is suboptimal to drop packets while simultaneously transmitting with low power or shutting off the wireless card. In order to make dropping packets suboptimal, it is necessary to penalize every dropped packet by at least as much as it would cost to admit it into the buffer. The largest cost that a packet can incur after being admitted into the buffer is approximately the infinite horizon holding cost that would be incurred if the packet is held in the buffer forever. Since a packet arriving in time slot $n_0$ does not incur any holding cost until time slot $n_0 + 1$, this infinite horizon holding cost can be computed as

$$\eta = \sum_{n=n_0+1}^{\infty} (\gamma)^{n-n_0} 1 = \frac{\gamma}{1-\gamma}, \qquad (13)$$

where $\gamma$ is the same discount factor used in (10) and (11), and $1$ is the instantaneous holding cost that is incurred by the packet in each time slot that it is held in the buffer.

## B. The Lagrangian approach

We can reformulate the constrained optimization in (12) as an unconstrained MDP by introducing a Lagrange multiplier associated with the delay constraint. We can then define a Lagrangian cost function:

$$c^\mu(s,a) = \rho(s,a) + \mu g(s,a), \qquad (14)$$

where $\mu \geq 0$ is the Lagrange multiplier, $\rho(s,a)$ is the power cost defined in (3), and $g(s,a)$ is the buffer cost defined in (8). From [28], we know that solving the constrained MDP problem is equivalent to solving the unconstrained MDP and its Lagrangian dual problem. We present this result in the following theorem.

**Theorem 1:** *The optimal value of the constrained MDP problem can be computed as*

$$L_\delta^{\pi^*,\mu^*}(s) = \min_{\pi \in \Phi} \max_{\mu \geq 0} V^{\pi,\mu}(s) - \mu\delta = \max_{\mu \geq 0} \min_{\pi \in \Phi} V^{\pi,\mu}(s) - \mu\delta, \qquad (15)$$

*where*

$$V^{\pi,\mu}(s) = E\left[\sum_{n=0}^{\infty} (\gamma)^n c^\mu\left(s^n, \pi(s^n)\right) \mid s^0 = s\right], \qquad (16)$$

*and a policy $\pi^*$ is optimal for the constrained MDP if and only if*

$$L_\delta^{\pi^*,\mu^*}(s) = \max_{\mu \geq 0} V^{\pi^*,\mu}(s) - \mu\delta. \qquad (17)$$

**Proof:** The detailed proof can be found in Chapter 3 of [28]. □

For a fixed $\mu$, the rightmost minimization in (15) is equivalent to solving the following dynamic programming equation:

$$V^{*,\mu}(s) = \min_{a \in \mathcal{A}} \left\{ c^\mu(s,a) + \gamma \sum_{s' \in \mathcal{S}} p(s' \mid s, a) V^{*,\mu}(s') \right\}, \forall s \in \mathcal{S} \qquad (18)$$

where $V^{*,\mu} : \mathcal{S} \mapsto \mathbb{R}$ is the *optimal state-value function*.

We also define the *optimal action-value function* $Q^{*,\mu} : \mathcal{S} \times \mathcal{A} \mapsto \mathbb{R}$, which satisfies:

$$Q^{*,\mu}(s,a) = c^\mu(s,a) + \gamma \sum_{s' \in \mathcal{S}} p(s' \mid s, a) V^{*,\mu}(s'), \qquad (19)$$

where $V^{*,\mu}(s') = \min_{a \in \mathcal{A}} Q^{*,\mu}(s', a)$. In words, $Q^{*,\mu}(s,a)$ is the infinite horizon discounted cost achieved by taking action $a$ in state $s$ and then following the optimal policy $\pi^{*,\mu}$ thereafter, where

$$\pi^{*,\mu}(s) = \arg\min_{a \in \mathcal{A}} Q^{*,\mu}(s,a), \ \forall s \in \mathcal{S}. \tag{20}$$

Assuming that the cost and transition probability functions are *known* and *stationary*, (18), (19), and (20) can be computed numerically using the well-known *value iteration* algorithm [11].

For notational simplicity, we drop the Lagrange multiplier from the notation in the remainder of the report unless it is necessary; hence, we will write $c(s,a)$, $V^*(s)$, $Q^*(s,a)$, $\pi^*(s)$ instead of $c^\mu(s,a)$, $V^{*,\mu}(s)$, $Q^{*,\mu}(s,a)$, and $\pi^{*,\mu}(s)$, respectively. We discuss how to learn the optimal $\mu$ in Section V.B. The relationships between the state and action at time $n$, the state at time $n+1$, the transition probability and cost functions, and the state-value function, are illustrated in Figure 2.

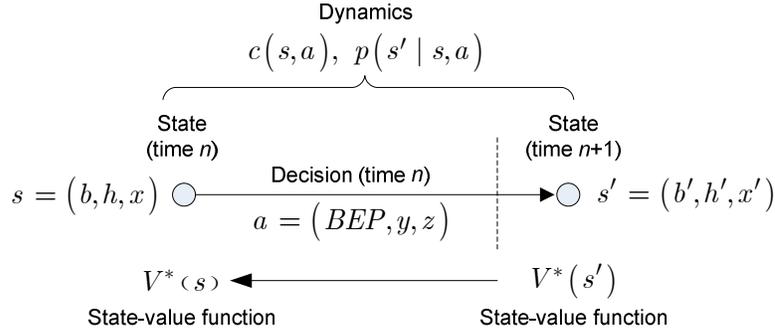

**Figure 2. Visualization of the MDP.**

## V. LEARNING THE OPTIMAL POLICY

In practice, the cost and transition probability functions are (partially) *unknown* a priori. Consequently, $V^*$ and $\pi^*$ cannot be computed using value iteration; instead, they must be learned online, based on experience. To this end, we adopt a *model-free* RL approach, which can be used to learn $Q^*$ and $\pi^*$ online, without first estimating the unknown cost and transition probability functions.

We begin in Section V.A by introducing a well-known model-free RL algorithm called *Q-learning*, which directly estimates $Q^*$ under the assumption that the system's dynamics are completely unknown a priori. In Section V.B, we develop a framework that allows us to integrate known information about the cost and transition probability functions into the learning process. Exploiting this partially known information dramatically improves run-time performance compared to the Q-learning algorithm.

## A. Conventional Q-learning

Central to the conventional Q-learning algorithm is a simple update step performed at the end of each time slot based on the *experience tuple* (ET) $\sigma^n = (s^n, a^n, c^n, s^{n+1})$: i.e.,

$$Q^{n+1}(s^n, a^n) \leftarrow (1 - \alpha^n) Q^n(s^n, a^n) + \alpha^n \left[ c^n + \gamma \min_{a' \in \mathcal{A}} Q^n(s^{n+1}, a') \right], \quad (21)$$

where $s^n$ and $a^n$ are the state and performed action in time slot $n$, respectively; $c^n$ is the corresponding cost with expectation $c(s^n, a^n)$; $s^{n+1}$ is the resulting state in time slot $n+1$, which is distributed according to $p(s^{n+1} \mid s^n, a^n)$; $a'$ is the *greedy action* in state $s^{n+1}$, which minimizes the current estimate of the action-value function; $\alpha^n \in [0,1]$ is a time-varying learning rate parameter; and, $Q^0(s, a)$ can be initialized arbitrarily for all $(s, a) \in \mathcal{S} \times \mathcal{A}$.

Q-learning essentially uses a sample average of the action-value function, i.e. $Q^n$, to approximate $Q^*$. It is well known that if (i) the instantaneous cost and transition probability functions are stationary, (ii) all of the state-action pairs are visited infinitely often[8], and (iii) $\alpha^n$ satisfies the *stochastic approximation conditions* $\sum_{n=0}^{\infty} \alpha^n = \infty$ and $\sum_{n=0}^{\infty} (\alpha^n)^2 < \infty$, then $Q^n$ converges with probability 1 to $Q^*$ as $n \to \infty$ (for example, $\alpha^n = 1/(n+1)$ or $\alpha^n = (1/n)^{0.7}$ satisfy the stochastic approximation conditions).[9] Subsequently, the optimal policy can be calculated using (20).

Using Q-learning, it is not obvious what the best action is to take in each state during the learning process. On the one hand, $Q^*$ can be learned by randomly *exploring* the available actions in each state. Unfortunately, unguided randomized exploration cannot guarantee acceptable run-time performance because suboptimal actions will be taken frequently. On the other hand, taking greedy actions, which *exploit* the available information in $Q^n$, can guarantee a certain level of performance, but exploiting what is already known about the system prevents the discovery of better actions. Many techniques are available in the literature to judiciously trade off exploration and exploitation. In this report, we use the so-called $\varepsilon$ -*greedy* action selection method [11], but other techniques such as Boltzmann exploration can also be deployed [11]. Importantly, the only reason why exploration is required is because Q-learning assumes that the unknown cost and transition probability functions depend on the action. In the remainder of this section, we will describe circumstances under which exploiting partial information about the system can obviate the need for action exploration.

---

[8] A state-action pair $(s, a)$ is said to be visited infinitely often if $\lim_{N \to \infty} \sum_{n=0}^{N} I(s^n = s \text{ and } a^n = a) = \infty$. In other words, if the system is allowed to run forever, then $(s, a)$ will be sampled an infinite number of times.

[9] Condition (i) will only hold if the arrival distribution and channel transition probabilities are stationary, which is often not true in practice. In situations where the dynamics are time-varying, faster learning algorithms like those proposed in Section V.B and V.C are required to closely track the dynamics and achieve near optimal performance. Condition (ii) is satisfied under an appropriate exploration policy [11] as long as the MDP is communicating [27]. Condition (iii) is trivially satisfied because we choose the sequence of learning rates $\{\alpha^n\}$.

Q-learning is an inefficient learning algorithm because it only updates the action-value function for a single state-action pair in each time slot and does not exploit known information about the system's dynamics. Consequently, it takes a long time for the algorithm to converge to a near-optimal policy, so run-time performance suffers [17] [18]. In the remainder of this section, we introduce two techniques that improve learning performance by exploiting known information about the system's dynamics and enabling more efficient use of the experience in each time slot.

## B. *Proposed post-decision state learning*

*1. Partially known dynamics*

The conventional Q-learning algorithm learns the value of state-action pairs under the assumption that the environment's dynamics (i.e. the cost and transition probability functions) are completely unknown a priori. In many systems, however, we have partial knowledge about the environment's dynamics. Table 3 highlights what is assumed known, what is assumed unknown, what is stochastic, and what is deterministic in this report. Note that this is just an illustrative example and the dynamics may be classified differently under different conditions. For example, if the parameter $\theta$ in (4) is known and in the interval $(0,1)$, then the power management state transition can be classified as stochastic and known as in [3]; if $\theta$ is unknown, then the power management state transition can be classified as stochastic and unknown; if the BEP function defined in (1) is unknown, then the goodput distribution and holding cost can be classified as stochastic and unknown; or, if the transmission power function in (2) is unknown, then the power cost can be classified as unknown. Importantly, the proposed framework can be applied regardless of the specific classification. In Section V.B.2, after introducing the PDS, we discuss how our PDS is a generalization of the PDS defined in [12], which can only be applied for specific classifications.

We will exploit the known information about the dynamics to develop more efficient learning algorithms than Q-learning, but first we need to formalize the concepts of known and unknown dynamics in our MDP-based framework.

**Table 3. Classification of dynamics.**

|  | Known | Unknown |
|---|---|---|
| **Deterministic** | ● Power management state transition [(4), $\theta = 1$]<br>● Power cost (3) | N/A |
| **Stochastic** | ● Goodput distribution (6) (Section III)<br>● Holding cost (8) | ● Traffic arrival distribution $p^l(l)$<br>● Channel state distribution $p^h(h' \mid h)$<br>● Overflow cost (8) |

*2. Generalized post-decision state definition*

We define a *post-decision state* (PDS) to describe the state of the system *after* the known dynamics take place, but *before* the unknown dynamics take place. As noted in the introduction, the PDS is a known construct in prior literature, see, e.g., [11] [12] [26], but we generalize the concept so that it can apply to the considered problem. We denote the PDS as $\tilde{s} \in \mathcal{S}$ (note that the set of possible PDSs is the same as the set of possible states). Based on the discussion in the previous subsection, the PDS at time $n$ is related to the state $s^n = (b^n, h^n, x^n)$, action $a^n = (BEP^n, y^n, z^n)$, and the state at time $n+1$, as follows:

- **PDS at time** $n$: $\tilde{s}^n = (\tilde{b}^n, \tilde{h}^n, \tilde{x}^n) = ([b^n - f^n], h^n, x^{n+1})$.
- **State at time** $n+1$: $s^{n+1} = (b^{n+1}, h^{n+1}, x^{n+1}) = ([b^n - f^n] + l^n, h^{n+1}, x^{n+1})$.

The buffer's PDS $\tilde{b}^n = b^n - f^n$ characterizes the buffer state *after* the packets are transmitted, but before new packets arrive; the channel's PDS is the same as the channel state at time $n$; and, the power management PDS is the same as the power management state at time $n+1$. In other words, the PDS incorporates all of the known information about the transition from state $s^n$ to state $s^{n+1}$ after taking action $a^n$. Meanwhile, the next state incorporates all of the unknown dynamics that were not included in the PDS (i.e. the number of packet arrivals $l^n$ and next channel state $h^{n+1}$). Note that the buffer state at time $n+1$ can be rewritten in terms of the buffer's PDS at time $n$ as $b^{n+1} = \tilde{b}^n + l^n$.

These relationships are illustrated in Figure 3, where it is also shown that several state-action pairs can lead to the same PDS. A consequence of this is that acquiring information about a *single* PDS provides information about the *many* state-action pairs that can potentially precede it. Indeed, this is the foundation of PDS-based learning, which we describe in Section V.B.4.

By introducing the PDS, we can factor the transition probability function into known and unknown components, where the known component accounts for the transition from the current state to the PDS, i.e. $s \to \tilde{s}$, and the unknown component accounts for the transition from the PDS to the next state, i.e. $\tilde{s} \to s'$. Formally,

$$p(s' \mid s,a) = \sum_{\tilde{s}} p_{\mathrm{u}}(s' \mid \tilde{s},a) p_{\mathrm{k}}(\tilde{s} \mid s,a), \qquad (22)$$

where the subscripts k and u denote the known and unknown components, respectively. We can factor the cost function similarly:

$$c(s,a) = c_{\mathrm{k}}(s,a) + \sum_{\tilde{s}} p_{\mathrm{k}}(\tilde{s} \mid s,a) c_{\mathrm{u}}(\tilde{s},a). \qquad (23)$$

The factorized transition probability and cost functions in (22) and (23), respectively, reduce to the factorized transition probability and cost functions that are implicitly used in [12] when $c_{\mathrm{u}}(\tilde{s},a) = 0$ and $p_{\mathrm{k}}(\tilde{s} \mid s^n, a)$ contains only 1s and 0s, i.e., when the transition from the state to the PDS is deterministic. Consequently, the algorithm in [12] cannot account for packet losses (because a deterministic PDS requires the goodput to be equal to the throughput) and cannot penalize buffer overflows (because all components of the cost function are assumed to be known, and the buffer overflow cost necessarily depends on the arrival distribution, which is unknown).

Notice that, in general, both the known and unknown components may depend on the action and the PDS. The proposed framework can be extended to this scenario, but action exploration will be necessary to learn the optimal policy. In the system under study, however, the unknown components only depend on the PDS (i.e. $p_\mathrm{u}(s' \mid \tilde{s}, a) = p_\mathrm{u}(s' \mid \tilde{s})$ and $c_\mathrm{u}(\tilde{s}, a) = c_\mathrm{u}(\tilde{s})$). The implication of this, discussed further in Section V.B.4, is that action exploration is not needed to learn the optimal policy. We will focus on the second scenario in the remainder of the report. Specifically, the known and unknown transition probability functions are defined as

$$p_\mathrm{k}(\tilde{s} \mid s, a) = p^x(\tilde{x} \mid x, y) p^f(b - \tilde{b} \mid BEP, z) I(\tilde{h} = h) \text{ and} \tag{24}$$

$$p_\mathrm{u}(s' \mid \tilde{s}) = p^h(h' \mid \tilde{h}) p^l(b' - \tilde{b}) I(x' = \tilde{x}), \tag{25}$$

where $I(\cdot)$ is the indicator function, which takes value 1 if its argument is true and 0 otherwise. Meanwhile, the known and unknown cost functions are defined as

$$c_\mathrm{k}(s, a) = \underbrace{\rho([h, x], BEP, y, z)}_{\text{power cost}} + \mu \sum_{f=0}^{z} p^f(f \mid BEP, z) \underbrace{[b - f]}_{\text{holding cost}}, \text{ and} \tag{26}$$

$$c_\mathrm{u}(\tilde{s}) = \mu \eta \sum_{l=0}^{\infty} p^l(l) \underbrace{\max(\tilde{b} + l - B, 0)}_{\text{overflow cost}}. \tag{27}$$

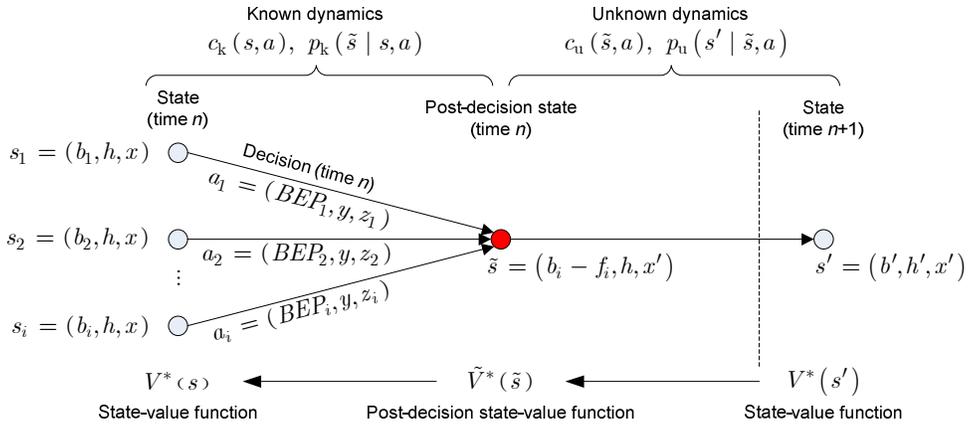

**Figure 3. Relationship between the state-action pair at time $n$, PDS at time $n$, and state at time $n+1$. State action pairs $(s_1, a_1), \ldots, (s_i, a_i)$ potentially lead to the same PDS.**

*3. Post-decision state-based dynamic programming*

Before we can describe the PDS learning algorithm, we need to define the *PDS value function*, which is a value function defined over the PDSs. In the PDS learning algorithm, the PDS value function plays a similar role as the action-value function does in the conventional Q-learning algorithm. The optimal PDS value function, denoted by $\tilde{V}^*$, can be expressed as a function of the optimal state-value function and vice-versa:

$$\tilde{V}^*(\tilde{s}) = c_\mathrm{u}(\tilde{s}) + \gamma \sum_{s'} p_\mathrm{u}(s' \mid \tilde{s}) V^*(s'), \tag{28}$$

$$V^*(s) = \min_{a \in \mathcal{A}} \left\{ c_\mathrm{k}(s, a) + \sum_{\tilde{s}} p_\mathrm{k}(\tilde{s} \mid s, a) \tilde{V}^*(\tilde{s}) \right\}. \tag{29}$$

Given the optimal PDS value function, the optimal policy can be computed as

$$\pi^*_{\mathrm{PDS}}(s) = \min_{a \in \mathcal{A}} \left\{ c_\mathrm{k}(s, a) + \sum_{\tilde{s}} p_\mathrm{k}(\tilde{s} \mid s, a) \tilde{V}^*(\tilde{s}) \right\}. \tag{30}$$

The following proposition proves that $\pi^*_{\mathrm{PDS}}$, defined in (30), and $\pi^*$, defined in (20), are equivalent.

**Proposition 1:** $\pi^*_{\mathrm{PDS}}$ *and* $\pi^*$ *are equivalent.*

**Proof:** To show that $\pi^*_{\text{PDS}}$ and $\pi^*$ are equivalent we substitute (28) into (30) as follows:

$$\begin{aligned}
\pi^*_{\text{PDS}}(s) &= \min_{a \in \mathcal{A}} \left\{ c_k(s,a) + \sum_{\tilde{s}} p_k(\tilde{s} \mid s, a) \tilde{V}^*(\tilde{s}) \right\} \\
&= \min_{a \in \mathcal{A}} \left\{ c_k(s,a) + \sum_{\tilde{s}} p_k(\tilde{s} \mid s, a) \left[ c_u(\tilde{s}) + \gamma \sum_{s'} p_u(s' \mid \tilde{s}) V^*(s') \right] \right\} \\
&= \min_{a \in \mathcal{A}} \left\{ c_k(s,a) + \sum_{\tilde{s}} p_k(\tilde{s} \mid s, a) c_u(\tilde{s}) + \gamma \sum_{\tilde{s}} \sum_{s'} p_k(\tilde{s} \mid s, a) p_u(s' \mid \tilde{s}) V^*(s') \right\} \\
&= \min_{a \in \mathcal{A}} \left\{ c(s,a) + \gamma \sum_{s'} p(s' \mid s, a) V^*(s') \right\} \\
&= \min_{a \in \mathcal{A}} \left\{ Q^*(s,a) \right\} \\
&= \pi^*(s).
\end{aligned}$$

where the fourth equality follows from (22) and (23). □

Proposition 1 is important because it enables us to use the PDS value function to learn the optimal policy.

*4. The post-decision state learning algorithm*

While Q-learning uses a sample average of the action-value function to approximate $Q^*$, PDS learning uses a sample average of the PDS value function to approximate $\tilde{V}^*$. However, because the value function $V$ can be directly computed from the PDS value function $\tilde{V}$ using the known dynamics and (29), the PDS learning algorithm *only needs to learn the unknown dynamics* in order to learn the optimal value function $V^*$ and the optimal policy $\pi^*$. The PDS learning algorithm is summarized in Table 4. The following theorem shows that the PDS learning algorithm converges to the optimal PDS value function $\tilde{V}^{*,\mu}(\tilde{s})$ in stationary environments.

**Theorem 2:** *The post-decision state learning algorithm converges to the optimal post-decision state value function $\tilde{V}^{*,\mu}(\tilde{s})$ in stationary environments when the sequence of learning rates $\alpha^n$ satisfies $\sum_{n=0}^{\infty} \alpha^n = \infty$ and $\sum_{n=0}^{\infty} (\alpha^n)^2 < \infty$.*

**Proof:** The post-decision state learning algorithm defined in Table 4 can be written using the recursion

$$\tilde{V}^{n+1}(\tilde{s}) \leftarrow \tilde{V}^n(\tilde{s}) + \alpha^n \left[ c_u(\tilde{s}) + \gamma \min_a \left\{ \begin{array}{l} c_k(\Psi^{n+1}(\tilde{s}), a) + \\ \sum_{\tilde{s}'} p_k(\tilde{s}' \mid \Psi^{n+1}(\tilde{s}), a) \tilde{V}^n(\tilde{s}') \end{array} \right\} - \tilde{V}^n(\tilde{s}) \right]$$

where $\tilde{s}, \tilde{s}' \in \mathcal{S}$, $a \in \mathcal{A}$, and $\Psi^{n+1}(\tilde{s}) \in \mathcal{S}$ is an independently simulated random variable drawn from $p_{\mathrm{u}}(\cdot \mid \tilde{s})$.

Let $h : \mathbb{R}^{|\mathcal{S}|} \to \mathbb{R}^{|\mathcal{S}|}$ be a map such that $h(\tilde{V}) = \left[h_{\tilde{s}}(\tilde{V})\right]_{\tilde{s}}$ with

$$h_{\tilde{s}}(\tilde{V}) = c_{\mathrm{u}}(\tilde{s}) + \gamma \sum_{s'} p_{\mathrm{u}}(s' \mid \tilde{s}) \min_{a} \left\{ c_{\mathrm{k}}(s', a) + \sum_{\tilde{s}'} p_{\mathrm{k}}(\tilde{s}' \mid s', a) \tilde{V}^n(\tilde{s}') \right\} - \tilde{V}^n(\tilde{s}),$$

where $\tilde{s}, \tilde{s}' \in \mathcal{S}$, $a \in \mathcal{A}$. Define $F(\tilde{V}) = \left[F_{\tilde{s}}(\tilde{V})\right]_{\tilde{s}}$ with

$$F_{\tilde{s}}(\tilde{V}) = c_{\mathrm{u}}(\tilde{s}) + \gamma \sum_{s'} p_{\mathrm{u}}(s' \mid \tilde{s}) \min_{a} \left\{ c_{\mathrm{k}}(s', a) + \sum_{\tilde{s}'} p_{\mathrm{k}}(\tilde{s}' \mid s', a) \tilde{V}^n(\tilde{s}') \right\}.$$

Then, $h(\tilde{V}) = F(\tilde{V}) - \tilde{V}$. As in [24], it can be shown that the convergence of the online learning algorithm is equivalent to the convergence of the associated O.D.E.

$$\dot{\tilde{V}} = F(\tilde{V}) - \tilde{V} := h(\tilde{V}).$$

Since the map $F : \mathbb{R}^{|\mathcal{S}|} \to \mathbb{R}^{|\mathcal{S}|}$ is a maximum norm $\gamma$-contraction [25], the asymptotic stability of the unique equilibrium point of the above O.D.E. is guaranteed [24]. This unique equilibrium point corresponds to the optimal post-decision state value function $\tilde{V}^{*,\mu}$. □

**Table 4. Post-decision state-based learning algorithm.**

| | |
|---|---|
| 1. | **Initialize:** At time $n = 0$, initialize the PDS value function $\tilde{V}^0$ as described in Section VI.E. |
| 2. | **Take the greedy action:** At time $n$, take the greedy action $$a^n = \arg\min_{a \in \mathcal{A}} \left\{ c_{\mathrm{k}}(s^n, a) + \sum_{\tilde{s}} p_{\mathrm{k}}(\tilde{s} \mid s^n, a) \tilde{V}^n(\tilde{s}) \right\}. \quad (31)$$ |
| 3. | **Observe experience:** Observe the PDS experience tuple $\tilde{\sigma}^n = \left(s^n, a^n, \tilde{s}^n, c_{\mathrm{u}}^n, s^{n+1}\right)$. |
| 4. | **Evaluate the state-value function:** Compute the value of state $s^{n+1}$: $$V^n(s^{n+1}) = \min_{a \in \mathcal{A}} \left\{ c_{\mathrm{k}}(s^{n+1}, a) + \sum_{\tilde{s}} p_{\mathrm{k}}(\tilde{s} \mid s^{n+1}, a) \tilde{V}^n(\tilde{s}) \right\}. \quad (32)$$ |
| 5. | **Update the PDS value function:** At time $n$, update the PDS value function using the information from steps 3 and 4: $$\tilde{V}^{n+1}(\tilde{s}^n) \leftarrow (1 - \alpha^n) \tilde{V}^n(\tilde{s}^n) + \alpha^n \left[c_{\mathrm{u}}^n + \gamma V^n(s^{n+1})\right] \quad (33)$$ |
| 6. | **Lagrange multiplier update:** Update the Lagrange multiplier $\mu$ using (34). |
| 7. | **Repeat:** Update the time index, i.e. $n \leftarrow n + 1$. Go to step 2. |

The optimal value of the Lagrange multiplier $\mu$ in (14), which depends on the delay constraint $\delta$, can be learned online using stochastic subgradients as in [12]: i.e.,

$$\mu^{n+1} = \Lambda[\mu^n + \beta^n (g^n - (1-\gamma)\delta)], \tag{34}$$

where $\Lambda$ projects $\mu$ onto $[0, \mu_{\max}]$, $\beta^n$ is a time-varying learning rate with the same properties as $\alpha^n$, $g^n$ is the buffer cost with expectation $g(s^n, a^n)$, and the $(1-\gamma)\delta$ term converts the discounted delay constraint $\delta$ to an average delay constraint. The following additional conditions must be satisfied by $\beta^n$ and $\alpha^n$ to ensure convergence of (34) to $\mu^*$ in Theorem 1:

$$\sum_{n=0}^{\infty} (\alpha^n + \beta^n) < \infty \text{ and } \lim_{n \to \infty} \frac{\beta^n}{\alpha^n} \to 0. \tag{35}$$

Although theoretical convergence of the PDS learning algorithm requires a stationary Markovian environment, it is possible to track non-stationary Markovian environmental dynamics by making a simple modification to the sequences of learning rates $\{\alpha^n\}$ (for the PDS value function update) and $\{\beta^n\}$ (for the Lagrange multiplier update). Specifically, keeping $\alpha^n$ and $\beta^n$ bounded away from zero will prevent past experience from heavily biasing the PDS value function and Lagrange multiplier, thereby allowing them to track the non-stationary dynamics. This strategy, when coupled with learning algorithms like PDS learning and virtual experience learning (see Section V.C), which can quickly converge in stationary Markovian environments (see our simulation results in Section VI.B), can achieve good system performance even in non-stationary Markovian environments (see our simulation results in Section VI.D).

There are several ways in which the PDS learning algorithm uses information more efficiently than Q-learning. First, PDS learning exploits partial information about the system so that less information needs to be learned. This is evident from the use of the known information (i.e. $c_{\mathrm{k}}(s^n, a)$ and $p_{\mathrm{k}}(\tilde{s}^n \mid s^n, a)$) in (31) and (32). Second, updating a single PDS using (33) provides information about the state-value function at many states. This is evident from the expected PDS value function on the right-hand-side of (32). Third, because the unknown dynamics (i.e. $c_{\mathrm{u}}(\tilde{s}^n)$ and $p_{\mathrm{u}}(s^{n+1} \mid \tilde{s}^n)$) do not depend on the action, there is no need for randomized exploration to find the optimal action in each state. This means that the latest estimate of the value function can always be exploited and therefore non-greedy actions never have to be tested.

## C. Proposed virtual experience learning

The PDS learning algorithm described in the previous subsection only updates one PDS in each time slot. In this subsection, we discuss how to update multiple PDSs in each time slot in order to accelerate the learning rate and improve run-time performance. The key idea is to realize that the traffic arrival and channel state distributions are independent of the buffer and power management states. Consequently, the statistical information obtained from the PDS at time $n$ about the a priori unknown traffic arrival and channel state distributions can be extrapolated to other PDSs with different post-decision buffer states and different post-decision power management states. Specifically, given the PDS experience tuple in time slot $n$, i.e. $\tilde{\sigma}^n = \left(s^n, a^n, \tilde{s}^n, c_{\mathrm{u}}^n, s^{n+1}\right)$, the PDS value function update defined in (33) can be applied to every *virtual experience tuple* in the set

$$\sum(\tilde{\sigma}^n) = \left\{\left(s^n, a^n, [\tilde{b}, \tilde{h}^n, \tilde{x}], c_{\mathrm{u}}(\tilde{b}; l^n), [\tilde{b} + l^n, h^{n+1}, \tilde{x}]\right) \mid \forall (\tilde{b}, \tilde{x}) \in \mathcal{B} \times \mathcal{X}\right\}, \qquad (36)$$

where $c_{\mathrm{u}}(\tilde{b}; l) = \eta \max(\tilde{b} + l - B, 0)$. We refer to elements of $\sum(\tilde{\sigma})$ as *virtual* experience tuples because they do not actually occur in time slot $n$. As shown in (36), the virtual experience tuples in $\sum(\tilde{\sigma}^n)$ are the same as the actual experience tuple $\tilde{\sigma}^n$ except that they have a different post-decision buffer state and/or a different post-decision power management state, and a different cost.

By performing the PDS update in (33) on every virtual experience tuple in $\sum(\tilde{\sigma}^n)$, the proposed virtual experience learning algorithm improves adaptation speed at the expense of a $|\mathcal{B} \times \mathcal{X}|$-fold increase in computational complexity ($|\mathcal{B} \times \mathcal{X}| = |\sum(\tilde{\sigma})|$). We show in the results section that we can trade off performance and learning complexity by only performing the complex virtual experience updates every $T$ time slots.

## D. Conceptual comparison of learning algorithms

Figure 4 illustrates the key differences between Q-learning (described in Section V.A), PDS learning (proposed in Section V.B), and virtual experience learning (proposed in Section V.C). To facilitate graphical illustration, Figure 4 uses a simplified version of the system model defined in Section III in which the state is fully characterized by the buffer state $b$, the only action is the throughput $z$, and there are no packet losses. Figure 4(a) illustrates how one Q-learning update on state-action pair $(b, z)$ only provides information about the buffer-throughput pair $(b, z)$. Figure 4(b) illustrates how one PDS learning update on PDS $b - z$ provides information about every state-action pair that can potentially lead to the PDS, which in this example corresponds to all buffer-throughput pairs $(b', z')$ such that $b' - z' = b - z$. Finally, Figure 4(c) illustrates how performing a PDS learning update on every virtual experience tuple associated with PDS $b - z$ provides information about every buffer state.

We now discuss the computational and memory complexity of the Q-learning, PDS learning, and virtual experience algorithms. But first, we note that in the considered system, the expectation on the right hand side of (31) and (32) can be efficiently computed as

$$\sum_{\tilde{s}} p_{\text{k}}\left(\tilde{s} \mid s^n, a\right) \tilde{V}^n\left(\tilde{s}\right) = \sum_{\substack{\tilde{x} \in \mathcal{X} \\ f \leq z}} p^x\left(\tilde{x} \mid x, y\right) p^f\left(f \mid BEP, z\right) \tilde{V}^n\left(b - f, h, \tilde{x}\right). \qquad (37)$$

The Q-learning algorithm described in Section V.A requires storing $|\mathcal{S}| \times |\mathcal{A}|$ action-values, i.e. $Q^n(s, a)$ for all $s \in \mathcal{S}$ and $a \in \mathcal{A}$. The PDS learning algorithm described in Section V.B requires storing $|\mathcal{S}|$ state-values, i.e. $V^n(s)$ for all $s \in \mathcal{S}$, and $|\mathcal{S}|$ post-decision state-values, i.e. $\tilde{V}^n(\tilde{s})$ for all $\tilde{s} \in \mathcal{S}$. Additionally, the PDS learning algorithm (implemented using (37)) requires storing the goodput distribution $p^f(f \mid BEP, z)$, power management transition probability function $p^x(x' \mid x, y)$, and the known cost function $c_{\text{k}}(s, a)$, which have sizes $|\mathcal{Z}|^2 |\mathcal{E}|$, $|\mathcal{X}|^2 |\mathcal{Y}|$, and $|\mathcal{S}| \times |\mathcal{A}|$ respectively. Table 6 shows that, using the state and action sets in our simulations (see Table 7 of Section VI), the PDS learning algorithm does not require much more memory than Q-learning, i.e. storage for 47,205 floating points compared to 45,760 floating points. Also note that PDS learning with virtual experience has the same memory requirements as PDS learning.

Table 5 illustrates both the greedy action selection complexity and learning update complexity for Q-learning, PDS learning, and virtual experience learning. It is clear from Table 5 that as more information is integrated into the learning algorithm, it becomes more computationally complex to implement. It is noteworthy that the virtual experience tuples in (37) can be updated in parallel using, for example, vector instructions. Hence, it is possible to implement virtual experience learning with as little complexity as PDS learning.

**Table 5.** Learning algorithm computational complexity (in each time slot). In the system under study, $\left|\tilde{\mathcal{S}}\right| = \left|\mathcal{X} \times \mathcal{Z}\right|$ and $\left|\Sigma\right| = \left|\mathcal{B} \times \mathcal{X}\right|$.

|  | Action Selection Complexity | Learning Update Complexity |
|---|---|---|
| Q-learning | $O(|\mathcal{A}|)$ | $O(|\mathcal{A}|)$ |
| PDS learning | $O(|\tilde{\mathcal{S}}||\mathcal{A}|)$ | $O(|\tilde{\mathcal{S}}||\mathcal{A}|)$ |
| Virtual experience learning | $O(|\tilde{\mathcal{S}}||\mathcal{A}|)$ | $O(|\Sigma||\tilde{\mathcal{S}}||\mathcal{A}|)$ |

**Table 6.** Learning algorithm memory complexity comparison using the parameters in Table 7 of Section VI.

| Set / Function | Cardinality / Memory complexity |
|---|---|
| States: $\mathcal{S} = \mathcal{B} \times \mathcal{H} \times \mathcal{X}$ | 26 x 8 x 2 = 416 |
| Actions: $\mathcal{A} = \mathcal{E} \times \mathcal{Y} \times \mathcal{Z}$ | 5 x 2 x 11 = 110 |
| Action-value function: $Q^n(s,a)$ | 416 x 110 = 45760 |
| State-value function: $V^n(s)$ | 416 |
| PDS value function: $\tilde{V}^n(\tilde{s})$ | 416 |
| Goodput distribution: $p^f(f \mid BEP, z)$ | 11 x 11 x 5 = 605 |
| Power management transition: $p^x(x' \mid x, y)$ | 2 x 2 x 2 = 8 |
| Known cost function: $c_k(s,a)$ | 416 x 110 = 45760 |
| **Q-learning total** | 45760 |
| **PDS learning / Virtual Experience total** | 45760 + 416 + 416 + 605 + 8 = 47205 |

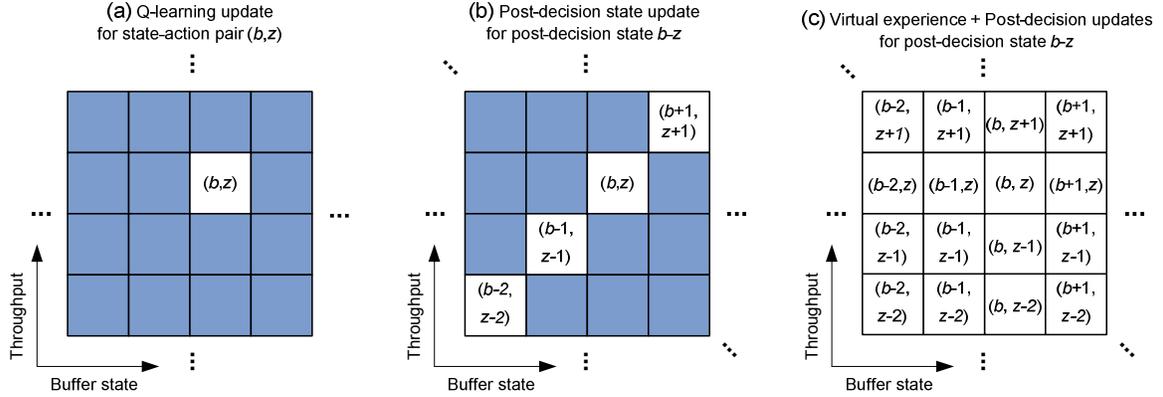

**Figure 4.** State-action pairs that are impacted in one time slot when using different learning updates (highlighted in white). (a) Q-learning update for state-action pair $(b, z)$; (b) PDS update for PDS $b - z$; (c) Virtual experience updates for PDS $b - z$.

*E. Initializing the PDS value function*

Given the known and unknown cost functions (i.e. $c_k(s,a)$ and $c_u(\tilde{s})$, respectively) and the known and unknown transition probability functions (i.e. $p_k(\tilde{s} \mid s, a)$ and $p_u(s' \mid \tilde{s})$, respectively), the optimal PDS value function can be computed using PDS value iteration as follows:

$$\tilde{V}_k(\tilde{s}) = c_u(\tilde{s}) + \gamma \sum_{s'} p_u(s' \mid \tilde{s}) V_k(s'), \text{ and} \quad (38)$$

$$V_{k+1}(s) = \min_{a \in \mathcal{A}} \left\{ c_k(s,a) + \sum_{\tilde{s}} p_k(\tilde{s} \mid s,a) \tilde{V}_k(\tilde{s}) \right\}, \tag{39}$$

where $k$ denotes the iteration index; $V_0(s)$ can be initialized arbitrarily; and $\tilde{V}_k$ converges to $\tilde{V}^*$ as $k \to \infty$. Of course, in practice, we do not know $c_u(\tilde{s})$ and $p_u(s' \mid \tilde{s})$. Hence, to initialize the PDS value function in step 1 of the PDS learning algorithm in Table 4, we can replace $c_u(\tilde{s})$ and $p_u(s' \mid \tilde{s})$ with reasonable estimates $\widehat{c_u}(\tilde{s})$ and $\widehat{p_u}(s' \mid \tilde{s})$, respectively, and then use PDS value iteration. Note that the estimates can be derived from past measurements of the traffic arrival and channel state distributions.

## VI. SIMULATION RESULTS

### A. Simulation Setup

Table 7 summarizes the parameters used in our MATLAB-based simulator. We assume that the PHY layer is designed to handle QAM rectangular constellations and that a Gray code is used to map the information bits into QAM symbols. The corresponding bit-error probability and transmission power functions can be found in [7]. We evaluate the performance of the proposed algorithm for several values of $P_{\text{on}}$: at 320 mW, $P_{\text{on}}$ is much larger than $P_{\text{tx}}$ as in typical 802.11a/b/g wireless cards [15]; at 80 mW, $P_{\text{on}}$ is closer to $P_{\text{tx}}$ as in cellular networks. The channel transition distribution $p^h(h' \mid h)$ and packet arrival distribution $p^l(l)$ are assumed to be unknown a priori in our simulations. Lastly, we choose a discount factor of $\gamma = 0.98$ in the optimization objective ($\gamma$ closer to 1 yields better performance after convergence, but requires more time to converge).

The signal-to-noise ratio (SNR) used for the tests is $SNR^n = \dfrac{P_{\text{tx}}^n}{N_0 W}$, where $P_{\text{tx}}^n$ is the transmission power in time slot $n$, $N_0$ is the noise power spectral density, and $W$ is the bandwidth. We assume that the bandwidth is equal to the symbol rate, i.e. $W = \dfrac{1}{T_s}$, where $T_s$ is the symbol duration.

### B. Learning algorithm comparison

Simulation results using the parameters described in Section VI.A and Table 7 are presented in Figure 5 for numerous simulations with duration 75,000 time slots (750 s) and $P_{\text{on}} = 320$ mW. Figure 5(a) illustrates the cumulative average cost versus time, where the cost is defined as in (14); Figure 5(b) illustrates the cumulative average power versus time, where the power is defined as in (3); Figure 5(c) and Figure 5(d) illustrate the cumulative average holding cost and cumulative average packet overflows versus time, respectively, as defined in (8); Figure 5(e) illustrates the cumulative average number of time slots spent in the off state (with both $x = \text{off}$ and $y = \text{s\_off}$), denoted by $\theta_{\text{off}}$, versus time; and Figure 5(f) illustrates the windowed average of the Lagrange multiplier versus time (with a window size of 1000 time slots).

**Table 7. Simulation Parameters.**

| Parameter | Value | Parameter | Value |
|---|---|---|---|
| Average arrival rate $\lambda$ | 200 packets/second | Packet loss rates $PLR$ | {1, 2, 4, 8, 16} % |
| Bits per symbol | {1, 2, ... , 10} | Packet size $\ell$ | 5000 bits |
| Buffer size $B$ (5) | 25 packets | Power management actions $y \in \mathcal{Y}$ (4) | {s\_on, s\_off} |
| Channel states $h \in \mathcal{H}$ | {-18.82, -13.79, -11.23, -9.37, -7.80, -6.30, -4.68, -2.08} dB | Power management states $x \in \mathcal{X}$ (4) | {on, off} |
| Discount factor $\gamma$ (16) | 0.98 | Symbol rate $1/T_s$ | $500 \times 10^3$ symbols/s |
| Holding cost constraint $(1-\gamma)\delta$ (12), (15), (34) | 4 packets | Time slot duration $\Delta t$ | 10 ms |
| Noise power spectral density $N_0$ | $2 \times 10^{-11}$ watts/Hz | Transition power $P_{\text{tr}}$ (3) | Set equal to $P_{\text{on}}$ |
| "Off" power $P_{\text{off}}$ (3) | 0 watts | Transmission actions $z \in \mathcal{Z}$ (3), (5), (8) | {0, 1, 2, ... , 10} packets/time slot |
| "On" power $P_{\text{on}}$ (3) | 80 mW, 160mW, or 320 mW | | |

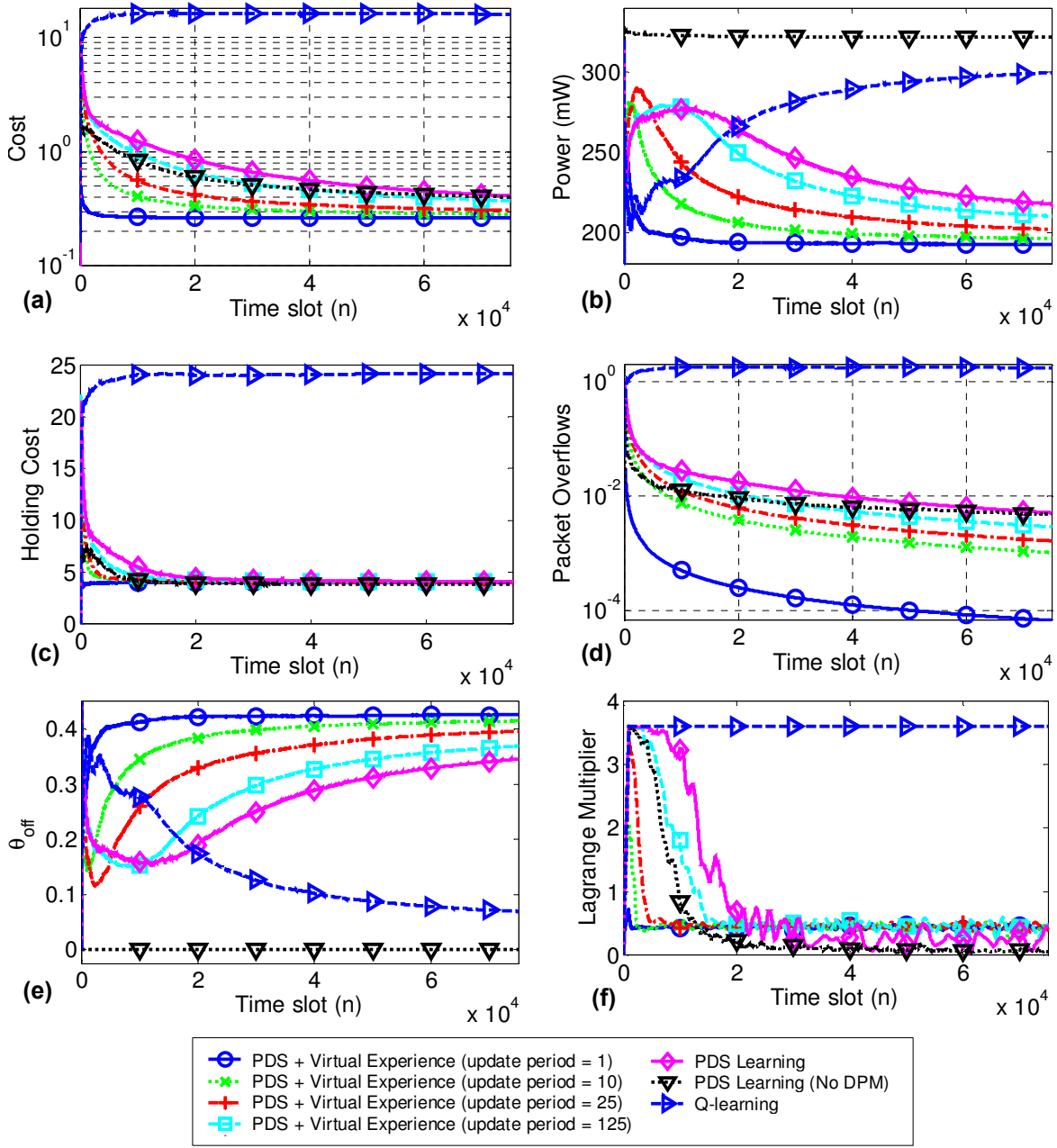

**Figure 5. Quantitative learning algorithm comparison with** $P_{\text{on}} = 320$ **mW. (a) Cumulative average cost vs. time slot. The y-axis is in log-scale due to the high dynamic range. (b) Cumulative average power vs. time slot. (c) Cumulative average holding cost vs. time slot. (d) Cumulative average packet overflows vs. time slot. The y-axis is in log-scale due to the high dynamic range. (e) Cumulative average** $\theta_{\text{off}}$ **vs. time slot. (f) Windowed average of Lagrange multiplier** $\mu$ **vs. time slot (window size = 1000 time slots).**

Each plot in Figure 5 compares the performance of the proposed PDS learning algorithm (with virtual experience updates every $T = 1, 10, 25, 125$ time slots) to the conventional Q-learning algorithm [11] and to an existing PDS learning-based solution [12]. The curves labeled "PDS Learning (No DPM)" are obtained by adapting the algorithm in [12] to use a discounted MDP formulation, rather than a time-average MDP formulation, and to include an adaptive BEP. This algorithm does not consider system-level power management (i.e. no DPM) and does not exploit virtual experience. Meanwhile, the curves labeled "PDS Learning" are obtained using the PDS learning algorithm proposed in Section V.B, which includes system-level power management.

In Figure 5, the PDS learning algorithms are initialized as described in Section V.E assuming that the arrival distribution is deterministic with 5 packet arrivals per time slot and that the channel transition matrix is the identity matrix.[10] Later, in Section VI.E, we discuss the sensitivity of the PDS learning algorithm to the initial PDS value function.

As expected, using a virtual experience update period of 1 leads to the best performance (see "PDS + Virtual Experience (update period = 1)" in Figure 5); in fact, this algorithm achieves approximately optimal performance after 3,000 time slots, at which point the cumulative average cost and Lagrange multiplier settle to nearly steady-state values (Figure 5(a) and Figure 5(f), respectively). Thus, if sufficient computational resources are available, the jointly optimal power-control, AMC, and DPM policies can be quickly learned online by smartly exploiting the structure of the problem using PDS learning combined with virtual experience. In existing wireless communications systems (especially in mobile devices), however, there may not be enough computational resources to use virtual experience in every time slot.[11] Hence, we also illustrate the performance of the proposed algorithm for different update periods (see "PDS + Virtual Experience (update period = 10, 25, 125)" in Figure 5) and without doing any virtual experience updates (see "PDS Learning" in Figure 5). As the number of updates in each time slot is decreased, the learning performance also decreases.

From the plots with different update periods, we can identify how the system adapts over time. Initially, the holding cost and packet overflows rapidly increase because the initial policy is not tuned to the specific traffic and channel conditions. This increasing delay drives the Lagrange multiplier to its predefined maximum value (resulting in a cost function that weighs delay more heavily and power less heavily), which leads to increased power consumption, which in turn drives the holding cost back down toward the holding cost constraint. As the cumulative average holding cost decreases towards the holding cost constraint, the Lagrange multiplier also begins to decrease, which eventually leads to a decrease in power consumption (because the cost function begins to weigh delay less heavily and power more heavily). Clearly, this entire process proceeds more quickly when there are more frequent virtual experience updates, thereby resulting in better overall performance. This process also explains why near optimal delay performance is achieved sooner than near optimal power consumption (because the minimum power consumption can only be learned *after* the holding cost constraint is satisfied).

---

[10] The channel transition matrix is defined from the channel state transition probability function as follows: $\mathbf{P}^h = \left[ p^h \left( h' \mid h \right) \right]_{h,h'}$.

[11] Here we assume that vector instructions are not available, so the update cannot be performed in parallel.

We now compare our proposed solution to the "PDS Learning (No DPM)" curves in Figure 5, which were obtained using the PDS learning algorithm introduced in [12] (modified slightly as described at the beginning of this subsection). It is clear that this algorithm performs approximately the same as our proposed "PDS + Virtual Experience (update period = 125)" algorithm in terms of delay (i.e. holding cost and overflows) but not in terms of power. The reason for the large disparity in power consumption is that the PDS learning algorithm introduced in [12] does not take advantage of system-level DPM, which is where the large majority of power savings comes from when $P_{on}$ is significantly larger than $P_{tx}$. This is corroborated by the fact that the $\theta_{off}$ curves in Figure 5(e) and the power consumption curves in Figure 5(b) are very similarly shaped. Thus, solutions that ignore system-level power management achieve severely suboptimal power in practice. (Note that PHY-centric power management solutions are still important for achieving the desired delay constraint!) Interestingly, despite using the same number of learning updates in each time slot, the "PDS Learning (No DPM)" algorithm performs better than "PDS Learning" algorithm in terms of delay (i.e. holding cost and overflow). This can be explained by the fact that DPM is ignored, so the learning problem is considerably simpler (i.e. there are less states and actions to learn about).

Lastly, we observe that the conventional Q-learning algorithm (see "Q-learning" in Figure 5) performs very poorly. Although Q-learning theoretically converges to the optimal policy, and therefore, should eventually approach the performance of the other algorithms, its cost in Figure 5(a) goes flat after its buffer overflows in Figure 5(d). It turns out that Q-learning has trouble finding the best actions to reduce the buffer occupancy because it requires action exploration and it only updates one state-action pair in each time slot. Nevertheless, it is clear from the increasing power cost in Figure 5(b) and the decreasing value of $\theta_{\text{off}}$ in Figure 5(e) that Q-learning is slowly learning the optimal policy: indeed, all of the learning algorithms must learn to keep the wireless card on before they can transmit enough packets to drain the buffer and satisfy the buffer constraint.

It is important to note that the learning algorithms converge to the optimal value and optimal policy faster than they appear to in Figure 5. The reason for this is that Figure 5 shows cumulative average results, which are biased by the initial suboptimal performance. To more accurately compare the convergence time scales of the various learning algorithms, we use the following *weighted percent error* metric, which is similar to the one proposed in Chapter 15.4.5 of [26]:

$$e^{n,\pi^*} = \sum_{s \in \mathcal{S}} p^*(s) \left| \frac{V^*(s) - V^n(s)}{V^*(s)} \right|,$$

where $p^*(s)$ is the stationary probability of being in state $s$ under the optimal policy $\pi^*$, $V^*(s)$ is the optimal state-value function, and $V^n(s)$ is the learned state-value function at the beginning of time slot $n$. We weight the estimation error $|V^*(s) - V^n(s)|$ by $p^*(s)$ because the optimal policy will often only visit a subset of the states, and the remaining states will never be visited; thus, if we were to estimate $V^*(s)$ while following the optimal policy $\pi^*$, a *nonweighted* estimation error would not converge to 0, but the weighted estimation error would. Using the weighted percent error metric, we define the $\epsilon$-convergence time as follows:

$$n^{\epsilon}_{\text{converge}} = \min\left\{ n \mid e^{t,\pi^*} < \epsilon, \text{ for all } t \geq n \right\}.$$

In words, the $\epsilon$-convergence time is the earliest time after which the weighted percent error is always below the threshold $\epsilon$.

Table **8** compares the $\epsilon$-convergence times of the various learning algorithms for $\epsilon = 0.1$. Note that virtual experience can converge up to two orders of magnitude faster than PDS learning without virtual

experience, and up to three orders of magnitude faster than Q-learning.

Table 8. convergence time for $\epsilon = 0.1$. Q-learning results are rounded up to the nearest 1000. All other results are rounded up to the nearest 100.

| Learning Algorithm | Convergence Time ($\epsilon = 0.1$) |
|---|---|
| PDS + Virtual Experience (update period = 1) | 100 |
| PDS + Virtual Experience (update period = 10) | 1500 |
| PDS + Virtual Experience (update period = 25) | 4800 |
| PDS + Virtual Experience (update period = 125) | 16200 |
| PDS Learning | 25100 |
| PDS Learning (No DPM) [12] | 24000 |
| Q-learning | 158000 |

*C. Comparison to a policy that is optimal with respect to past history*

In Figure 6, we compare the performance of the best learning algorithm in Figure 5 (i.e., "PDS + Virtual Experience (update period = 1)") to the performance of a learning algorithm that is optimal with respect to past experience. Specifically, in time slot $n$, for $n = 0,1,...$, the optimal learning algorithm uses value iteration [11] to compute a policy that is optimal with respect to estimates of the arrival distribution $p^l(l)$ and channel transition $p^h(h' \mid h)$. The estimates in time slot $n$, denoted by $\widehat{p^{n,l}}(l)$ and $\widehat{p^{n,h}}(h' \mid h)$, are determined from previous packet arrival and channel realizations using maximum likelihood estimation: i.e., $\widehat{p^{n,l}}(l) = \frac{1}{n}\sum_{t=0}^{n-1} I(l^t = l)$ and $\widehat{p^{n,h}}(h' \mid h) = \frac{1}{n}\sum_{t=0}^{n-1} I(h^{t+1} = h' \mid h^t = h)$, where $I(\cdot)$ is the indicator function. They are then substituted into the definitions of the buffer transition probability function in (7) and the buffer cost in (8), which, in turn, are substituted into the state transition probability function in (9) and the Lagrangian cost in (14). Finally, value iteration [11] is used to compute the optimal policy with respect to the estimated state transition probability and Lagrangian cost functions.

The optimal learning algorithm described above is the best that any learning algorithm can do given past experience. However, it is not practical to implement in a real system due to the complexity of value iteration. For this reason, we use it as a benchmark against which we can compare the PDS learning algorithm with virtual experience proposed in Section V.C.

It is clear from Figure 6 that the optimal learning algorithm initially achieves suboptimal power and delay performance. This is because it has limited samples of the environment (i.e. the packet arrival and channel transition distributions) with which to estimate the optimal policy. It is also clear that the proposed PDS learning algorithm with virtual experience initially performs worse than the optimal learning algorithm. This is because the proposed algorithm not only needs to sample the environment, but, due to complexity constraints, it also needs to incrementally learn the PDS value function as described in Table 4. Nevertheless, within approximately 200 (respectively, 3000) time slots, the proposed learning algorithm performs as well as the optimal learning algorithm in terms of the average holding cost (respectively, average power consumption).

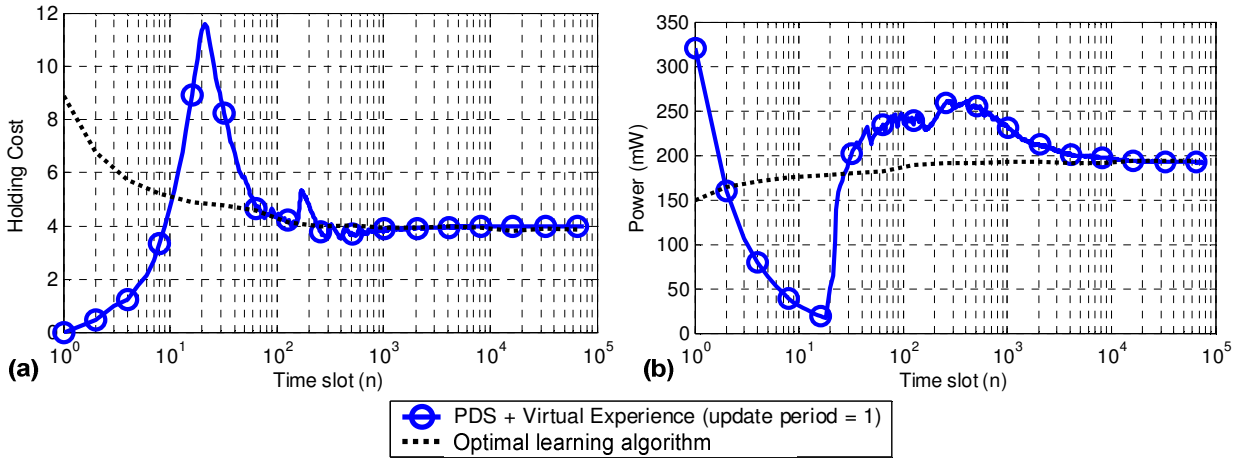

**Figure 6. Comparison of the optimal learning algorithm to the PDS learning algorithm with virtual experience and update period = 1. The x-axis is in log-scale to best highlight the learning process over time. (a) Holding cost vs. time slot. (b) Power vs. time slot.**

*D. Performance under non-stationary and non-Markovian dynamics*

In this subsection, we compare our proposed solution to the network energy saving solution in [6], which combines transmission scheduling with DPM, but uses a fixed BEP. Like our proposed solution, the DPM and packet scheduling solution in [6] uses a buffer to enable power-delay trade-offs. Unlike our proposed solution, however, the solution in [6] ignores the channel and traffic dynamics and is based on a simple *threshold-$k$ policy*: That is, if the buffer backlog exceeds $k$ packets, then the wireless card is turned on and all backlogged packets are transmitted as quickly as possible; then, after transmitting the packets, the wireless card is turned off again.

Figure 7 illustrates the average power-delay performance achieved by the proposed PDS learning algorithm (with virtual experience updates every $T=50$ time slots) and the threshold-$k$ policy after 75,000 time slots for $P_{\text{on}} \in \{80, 160, 320\}$ mW and $k = \{3, 5, 7, 9, 11, 13, 15\}$. All curves are generated assuming a fixed PLR of 1% and are obtained under non-stationary arrival dynamics[12] and channel dynamics[13]. The proposed algorithm performs between 10 and 30 mW better than the threshold-$k$ policy across the range of holding cost constraints and values of $k$ despite the fact that RL algorithms can only converge to optimal solutions in stationary and Markovian environments. Interestingly, the results in Figure 7 show that, for low values of $P_{\text{on}}$, the average power can actually *increase* as the delay increases under the threshold-$k$ policy. This is because, after the buffer backlog exceeds the predefined threshold $k$, the wireless card is turned on and transmits as many packets as possible regardless of the channel quality. Hence, a significant amount of transmit power is wasted in bad channel states. However, as $P_{\text{on}}$ increases, the results in Figure 7 show that the performance begins to improve and more closely approximate the optimal power-delay trade-off. Indeed, a more aggressive scheduling policy becomes necessary as $P_{\text{on}}$ increases so that the wireless card may be in the "off" state more frequently. The threshold-$k$ policy takes this observation to the extreme by transmitting the maximum number of packets possible in order to maximize the time spent in the "off" state. Unfortunately, because it ignores the transmission power, the threshold-$k$ policy cannot perform optimally, especially for smaller values of $P_{\text{on}}$.

---

[12] The arrival dynamics are generated by a 5-state Markov chain with states (0, 100, 200, 300, 400) packets/s and corresponding stationary distribution (0.0188, 0.3755, 0.0973, 0.4842, 0.0242). The state indicates the expected arrival rate for a Poisson arrival distribution. Importantly, the MDP does not know the traffic state; therefore, from its point of view, the traffic is non-stationary and non-Markovian.

[13] Non-stationary channel dynamics are generated by randomly perturbing the channel state transition probability $p(h' \mid h)$ from time slot to time slot.

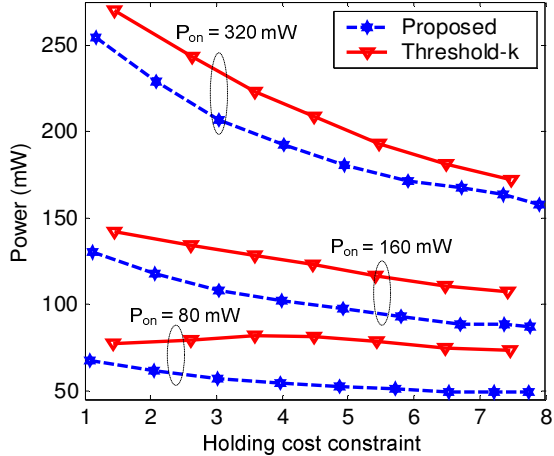

**Figure 7. Power-delay performance of proposed solution (with virtual experience updates every $T = 50$ time slots) compared to the power-delay performance of the threshold-$k$ policy for $k = \{3, 5, 7, 9, 11, 13, 15\}$. Traffic and channel dynamics are non-stationary.**

We now examine the sensitivity of the proposed solution to non-stationary and non-Markovian traffic, which we simulate using a variable bit-rate (VBR) video trace obtained by encoding 300 frames of the Foreman sequence using an H.264 video encoder (30 frames per second, CIF resolution, IBBPBB group of pictures structure, 1.7 Mbps encoded bitrate). It is important to note that the VBR traffic load varies rapidly (approximately every 33 ms) due to the different frame types in the group of pictures structure.[14] Additionally, the last 100 frames of the Foreman sequence generate much heavier traffic than the first 200 frames because of the fast motion induced by the panning camera. Overall, the dynamics are very challenging for any RL algorithm to adapt to.

Figure 8(a) illustrates the average power-delay performance of the proposed solution (with virtual experience updates in every time slot and the delay constraints shown on the x-axis) compared to the average performance of the threshold-$k$ policy for $k = \{3, 5, 7, 9, 11, 13, 15\}$. For large values of $k$, the threshold-$k$ policy outperforms the proposed solution due to the violation of the Markovian assumption. Although our solution performs well, it is not provably optimal for non-Markovian dynamics. This finding correlates with prior literature on dynamic power management [3], where it has been shown that simple timeout policies can occasionally outperform MDP-based policies in non-Markovian environments. Note that the threshold-$k$ policy performs poorly for lower values of $k$ (i.e., 3, 5, and 7) because it turns off the wireless card without anticipating future traffic arrivals, thereby forcing it to immediately turn the card back on and incur unnecessary power management transition power and delay overheads.

---

[14] In fact, the group of pictures structure itself can be modeled as a Markov chain by introducing additional states corresponding to the different frame types [35]. In this way, we could convert the non-Markovian VBR video trace to a Markovian VBR video trace. We could easily integrate these new states into the proposed framework, however, that is beyond the scope of the report. Instead, we focus on the non-Markovian VBR trace so that we can understand how the proposed algorithms perform when the Markovian assumption is violated.

Figure 8(b) illustrates, at a finer granularity than Figure 8(a), the sensitivity of the proposed solution (with virtual experience updates in every time slot and a delay constraint of 4 packets) to non-Markovian dynamics. Each point in Figure 8(b) corresponds to the average power (left y-axis) or delay (right y-axis) performance obtained in a simulation using an arrival process that is a convex combination of a stationary Poisson arrival process and the aforementioned VBR video trace, both with an average arrival rate of $340$ packets/s (1.7 Mbps). The convex combination is characterized by the parameter $\omega \in [0,1]$ shown on the x-axis of Figure 8(b). When $\omega = 0$, the traffic is stationary and Poisson; when $\omega = 1$ it is taken from the VBR video trace; and for $\omega \in (0,1)$ the traffic is stationary Poisson with probability $1-\omega$ and taken from the VBR trace with probability $\omega$. As $\omega$ increases, the traffic becomes increasingly non-Markovian, which causes the proposed algorithm to consume more power. Although the delay increases slightly as $\omega$ increases, it does not vary significantly from the delay constraint of 4 packets. In other words, even in rapidly time-varying non-Markovian environments, the proposed algorithm can successfully adapt to satisfy the application delay constraint.

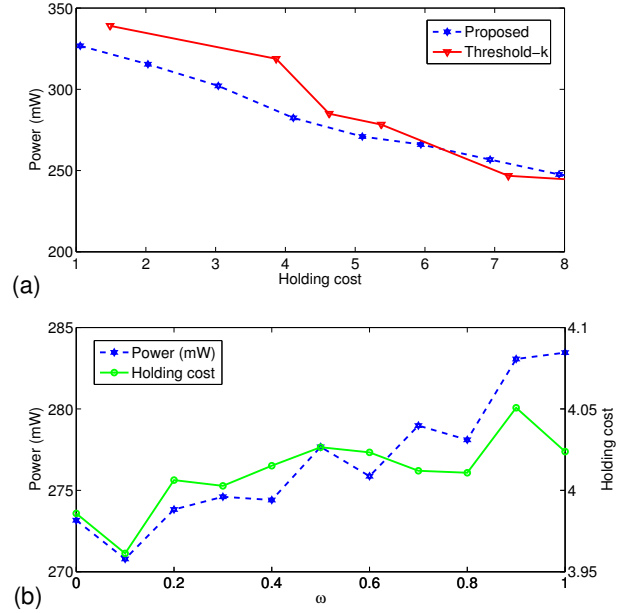

(a)

(b)

**Figure 8. Sensitivity to non-stationary and non-Markovian packet arrivals. (a) Average power-delay performance of proposed solution (with virtual experience updates in every time slot) compared to the average performance of the threshold-$k$ policy for $k = \{3,5,7,9,11,13,15\}$. (b) Comparison of power (left y-axis) and delay (right y-axis) performance of proposed solution for $\omega \in [0,1]$.**

## *E. Sensitivity to the initial PDS value function*

Figure 9 illustrates the sensitivity of the PDS learning algorithm to the initialized arrival rate which impacts $p_{\mathrm{u}}(s' \mid \tilde{s})$ through the buffer transition and impacts $c_{\mathrm{u}}(\tilde{s})$ through the overflow cost. In Figure 9, the operating points under stationary dynamics use the same arrival and channel dynamics as in Section VI.B and VI.C, and the operating points under non-stationary dynamics use the same arrival and channel dynamics as in Section VI.D. These empirical results demonstrate that initializing the traffic arrival distribution as deterministic leads to good performance that is relatively insensitive to the initialized arrival rate in both stationary and non-stationary environments. These results also demonstrate that poorly selecting the arrival distribution (e.g. assuming arrivals are uniformly distributed over $\{0,1,\ldots,B\}$) can adversely impact the power-delay performance. In all of the PDS learning results in this report, we assume that the channel transition matrix is initialized as the identity matrix.

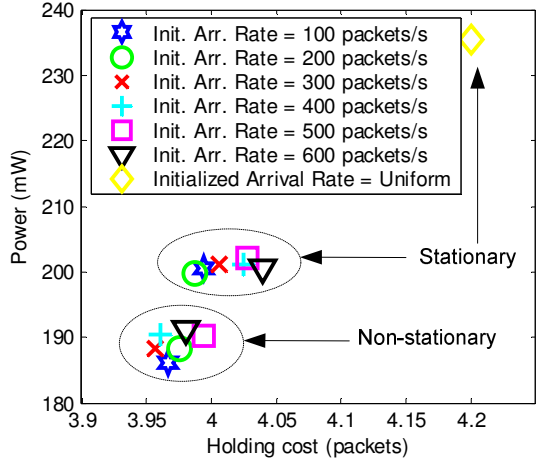

**Figure 9. Sensitivity to the arrival distribution used to initialize the PDS value function. The arrival distribution is assumed to be deterministic or uniformly distributed over $\{0,1,\ldots,B\}$. Virtual experience updates are used every $T$ = 25 time slots.**

*F. Summary of results*

The proposed virtual experience learning algorithm can be adapted to tradeoff implementation complexity and convergence time. It can converge up to two orders of magnitude faster than PDS learning without virtual experience, and up to three orders of magnitude faster than conventional Q-learning. Moreover, virtual experience learning can perform approximately as well as an optimal RL algorithm within approximately 200 time slots (for delay) and 3000 time slots (for power). Under slowly time-varying, non-stationary dynamics, the proposed algorithm achieves 11-33% improvement in power consumption for the same delay compared to the static threshold-$k$ policy. Under highly time-varying, non-stationary and non-Markovian dynamics, which are characteristic of variable bit-rate video traces, our solution performs well, and consistently satisfies the delay constraint; however, it is not provably optimal and occasionally performs worse than the static threshold-$k$ policy.

## VII. CONCLUSION

In this report, we considered the problem of energy-efficient point-to-point transmission of delay-sensitive multimedia data over a fading channel. We proposed a unified reinforcement learning solution for finding the jointly optimal power-control, AMC, and DPM policies when the traffic arrival and channel statistics are unknown. We exploited the structure of the problem by introducing a post-decision state, eliminating action-exploration, and enabling virtual experience to dramatically improve performance compared to conventional RL algorithms. Our experimental results demonstrate that the proposed solution outperforms existing solutions, even under nonstationary traffic and channel conditions. The results also strongly support the use of system-level power management solutions in conjunction with PHY-centric solutions to achieve the minimum possible power consumption, under delay constraints, in wireless communication systems.

Importantly, the proposed framework can be applied to any network or system resource management problem involving controlled buffers. One interesting future direction is to apply it in a system with multiple users (e,g. uplink or downlink transmission in cellular systems). This could be easily achieved by integrating our single-user optimization with one of the multi-user resource allocation frameworks proposed in, for example, [30] and [31].

## References

[1] D. Rajan, A. Sabharwal, and B. Aazhang, "Delay-bounded packet scheduling of bursty traffic over wireless channels," *IEEE Trans. on Information Theory*, vol. 50, no. 1, Jan. 2004.
[2] R. Berry and R. G. Gallager, "Communications over fading channels with delay constraints," *IEEE Trans. Inf. Theory*, vol 48, no. 5, pp. 1135-1149, May 2002.
[3] L. Benini, A. Bogliolo, G. A. Paleologo, and G. De Micheli, "Policy optimization for dynamic power management," IEEE Trans. on computer-aided design of integrated circuits, vol. 18, no. 6, June 1999.
[4] E.-Y. Chung, L. Benini, A. Bogliolo, Y.-H. Lu, and G. De Micheli, "Dynamic power management for nonstationary service requests," IEEE Trans. on Computers, vol. 51, no. 11, Nov. 2002.
[5] Z. Ren, B. H. Krogh, R. Marculescu, "Hierarchical adaptive dynamic power management," IEEE Trans. on Computers, vol. 54, no. 4, Apr. 2005.
[6] K. Nahrstedt, W. Yuan, S. Shah, Y. Xue, and K. Chen, "QoS support in multimedia wireless environments," in Multimedia Over IP and Wireless Networks, ed. M. van der Schaar and P. Chou, Academic Press, 2007.
[7] J. G. Proakis, *Digital Communications*. New York: McGraw-Hill, 2001.
[8] C. Schurgers, Energy-aware wireless communications. Ph.D. dissertation, University of California at Los Angeles, 2002.
[9] D. V. Djonin and V. Krishnamurthy, "MIMO transmission control in fading channels – a constrained Markov decision process formulation with monotone randomized policies," *IEEE Trans. on Signal Processing*, vol. 55, no. 10, Oct. 2007.
[10] D. Bertsekas, and R. Gallager, "Data networks," Prentice Hall, Inc., Upper Saddle River, NJ, 1987.
[11] R. S. Sutton and A. G. Barto, "Reinforcement learning: an introduction," Cambridge, MA:MIT press, 1998.
[12] N. Salodkar, A. Bhorkar, A. Karandikar, V. S. Borkar, "An on-line learning algorithm for energy efficient delay constrained scheduling over a fading channel," *IEEE Journal on Selected Areas in Communications*, vol. 26, no. 4, pp. 732-742, Apr. 2008.
[13] M. H. Ngo and V. Krishnamurthy, "Monotonocity of constrained optimal transmission policies in correlated fading channels with ARQ," *IEEE Trans. on Signal Processing*, vol. 58, No. 1, pp. 438-451, Jan. 2010.
[14] N. Mastronarde and M. van der Schaar, "Online reinforcement learning for dynamic multimedia systems," *IEEE Trans. on Image Processing*, vol. 19, no. 2, pp. 290-305, Feb. 2010.
[15] Cisco, Cisco Aironet 802.11a/b/g Wireless CardBus Adapter Data Sheet, Available online: http://www.cisco.com/en/US/prod/collateral/wireless/ps6442/ps4555/ps5818/product_data_sheet09186a00801ebc29.html

[16] F. Fu and M. van der Schaar, "Structural-aware stochastic control for transmission scheduling," Technical Report. Available online: http://medianetlab.ee.ucla.edu/papers/UCLATechReport_03_11_2010.pdf
[17] E. Evan-Dar and Y. Mansour, "Learning rates for Q-learning," *Journal of Machine Learning Research*, vol. 5, pp. 1-25, 2003.
[18] P. Auer, T. Jaksch, and R. Ortner, "Near-optimal regret bounds for reinforcement learning," *Advances in Neural Information Processing Systems*, vol. 21, pp. 89-96, 2009.
[19] Q. Liu, S. Zhou, and G. B. Giannakis, "Queuing with adaptive modulation and coding over wireless links: cross-layer analysis and design," *IEEE Trans. on Wireless Communications*, vol. 4, no. 3, pp. 1142-1153, May 2005.
[20] D. Krishnaswamy, "Network-assisted link adaptation with power control and channel reassignment in wireless networks," in *Proc. 3G Wireless Conf.*, 2002, pp. 165–170.
[21] K.-B. Song and S. A. Mujtaba, "On the code-diversity performance of bit-interleaved coded OFDM in frequency-selective fading channels," in *Proc. IEEE Veh. Technol. Conf.*, 2003, vol. 1, pp. 572–576.
[22] A. Goldsmith and S.-G. Chua, "Adaptive coded modulation for fading channels," *IEEE Trans. on Communications*, vol. 46, no. 5, pp. 595-602, May 1998.
[23] E. Altman, *Constrained Markov Decision Processes*. Boca Raton, FL: Chapman and Hall/CRC Press, 1999.
[24] V. S. Borkar, S. P. Meyn, "The ODE method for convergence of stochastic approximation and reinforcement learning," *SIAM J. Control Optim*, vol 38, pp. 447-469, 1999.
[25] D. P. Bertsekas, "Dynamic programming and optimal control," 3rd, Athena Scientific, Massachusetts, 2005.
[26] W. B. Powell, *Approximate Dynamic Programming: Solving the Curse of Dimensionality*, 2nd edition. New York: Wiley, 2011.
[27] M. Puterman, *Markov Decision Processes: Discrete Stochastic Dynamic Programming*. New York: Wiley, 1994.
[28] E. Altman, *Constrained Markov Decision Processes*. Chapman and Hall/CRC, 1999.
[29] V. S. Borkar, "An actor-critic algorithm for constrained Markov decision processes," *Systems & Control Letters*, vol. 54, pp. 207-2013, 2005.
[30] N. Mastronarde, F. Verde, D. Darsena, A. Scaglione, and M. van der Schaar, "Transmitting important bits and sailing high radio waves: a decentralized cross-layer approach to cooperative video transmission," to appear in *IEEE J. Sel. Areas Commun.* Available online: http://arxiv.org/abs/1102.5437v2
[31] N. Salodkar, A. Karandikar, V. S. Borkar, "A stable online algorithm for energy-efficient multiuser scheduling," *IEEE Trans. on Mobile Computing*, vol. 9, no. 10, pp. 1391-1406, Oct. 2010.
[32] C. Poellabauer and K. Schwan, "Energy-aware traffic shaping for wireless real-time applications," *Proc. of the 10th Real-Time and Embedded Technology and Applications Symposium*, pp. 48-55, May 2004.
[33] E. Shih, P. Bahl, and M. J. Sinclair, "Wake on Wireless: An Event Driven Energy Saving Strategy for Battery Operated Devices," *MOBICOM*, Sept. 23-26, 2002.
[34] N. Mastronarde and M. van der Schaar, "Fast reinforcement learning for energy-efficient wireless communications," Technical report. Available online: http://arxiv.org/abs/1009.5773
[35] D. S. Turaga and T. Chen, "Hierarchical modeling of variable bit rate video sources," *Packet Video Workshop*, May 2001.